\documentclass[10pt,twocolumn,letterpaper]{article}

\newif\ifarxiv
\arxivtrue

\ifarxiv
\usepackage[pagenumbers]{cvpr} %
\else
\usepackage{cvpr}              %
\fi

\definecolor{cvprblue}{rgb}{0.21,0.49,0.74}
\usepackage[pagebackref,breaklinks,colorlinks,allcolors=cvprblue]{hyperref}

\title{ETAP: Event-based Tracking of Any Point}

\author{Friedhelm Hamann$^{1, 4, 6}$, Daniel Gehrig$^{2}$, Filbert Febryanto$^{1, 4, 6}$,\\Kostas Daniilidis$^{2, 5}$, and Guillermo Gallego$^{1, 3, 4, 6}$.\\
$^{1}$~Technische Universit\"at Berlin, $^{2}$~University of Pennsylvania, $^{3}$~Einstein Center for Digital Future,\\
$^{4}$~Robotics Institute Germany, $^{5}$~Archimedes, Athena RC, $^{6}$~Science of Intelligence Excellence Cluster.
}

\usepackage{booktabs}
\usepackage{siunitx}  
\usepackage{multirow}
\usepackage{adjustbox} 
\usepackage{makecell}
\usepackage{pifont}
\usepackage{balance}
\usepackage[accsupp]{axessibility}
\usepackage[absolute]{textpos}

\newcommand{\gred}[1]{#1} %

\newcommand{\unum}[2]{\multicolumn{1}{c}{\underline{\tablenum[table-format={#1}]{#2}}}}
\newcommand{\bnum}[1]{\bfseries #1}
\newcommand{\subsubsec}[1]{%
    \noindent\textbf{#1}%
}

\def\cE{\mathcal{E}} %
\def\cD{\mathcal{D}} %
\def\pol{p} %
\def\bx{\mathbf{x}}
\def\cP{\mathcal{P}} %
\def\Real{\mathbb{R}} %
\def\Lum{L} %

\newcommand{\cmark}{\ding{51}}%
\newcommand{\xmark}{\ding{55}}%

\usepackage{xcolor}
\definecolor{light-gray}{gray}{0.6}
\newcommand\gframe[1]{{\color{light-gray}\frame{#1}}}

\newcommand{\dname}{EventKubric}  %
\newcommand{\mname}{ETAP}  %
\newcommand{\lname}{feature alignment}  %

\begin{document}

\ifarxiv
\definecolor{somegray}{gray}{0.5}
\newcommand{\darkgrayed}[1]{\textcolor{somegray}{#1}}
\begin{textblock}{11}(2.3, 0.8)  %
\begin{center}
\darkgrayed{This paper has been accepted for publication at the\\
IEEE Conference on Computer Vision and Pattern Recognition (CVPR), Nashville, 2025.
\copyright IEEE}
\end{center}
\end{textblock}
\fi

\maketitle
\begin{abstract}
Tracking any point (TAP) recently shifted the motion estimation paradigm from focusing on individual salient points with local templates to tracking arbitrary points with global image contexts.
However, while research has mostly focused on driving the accuracy of models in nominal settings, addressing scenarios with difficult lighting conditions and high-speed motions remains out of reach due to the limitations of the sensor.
This work addresses this challenge with the first event camera-based TAP method. 
It leverages the high temporal resolution and high dynamic range of event cameras for robust high-speed tracking, and the global contexts in TAP methods to handle asynchronous and sparse event measurements. 
We further extend the TAP framework to handle event feature variations induced by motion --- thereby addressing an open challenge in purely event-based tracking --- with a novel {\lname}-loss which ensures the learning of motion-robust features.
Our method is trained with data from a new data generation pipeline and systematically ablated across all design decisions.
Our method shows strong cross-dataset generalization and performs \gred{136}\% better on the average Jaccard metric than the baselines.
Moreover, on an established feature tracking benchmark, it achieves a \gred{20}\% improvement over the previous best event-only method and even surpasses the previous best events-and-frames method by \gred{4.1}\%.
Our code is available at \url{https://github.com/tub-rip/ETAP}.
\end{abstract}    
\section{Introduction}
\label{sec:intro}

Understanding scene motion from a video remains a fundamental challenge in computer vision, with renewed interest through its formulation as tracking any point (TAP)~\cite{Sand08ijcv,Harley22eccv,Doersch22neurips}.
A new class of powerful methods has been quickly adopted for downstream tasks, e.g., in robotics~\cite{Vecerik24icra, Chen24cvpr}.
However, existing methods focus on tracking in nominal settings due to the fundamental limitations of the sensor.

Event cameras represent a novel class of visual sensors offering high temporal resolution, high dynamic range (HDR), and low power consumption, characteristics that make them valuable stand-alone sensors for various robotic perception tasks.
These innovative sensors address several limitations of conventional cameras, particularly handling motion blur and high-speed movements.
To fully leverage these advantages we focus on developing an event-only method for tracking arbitrary 2D points within the data stream without additional sensor input.

\def\figmethodwidth{.48\linewidth}
\begin{figure}[t]
   \centering
   {\includegraphics[trim={0cm 3cm 2.3cm 1.2cm},clip,width=\linewidth]{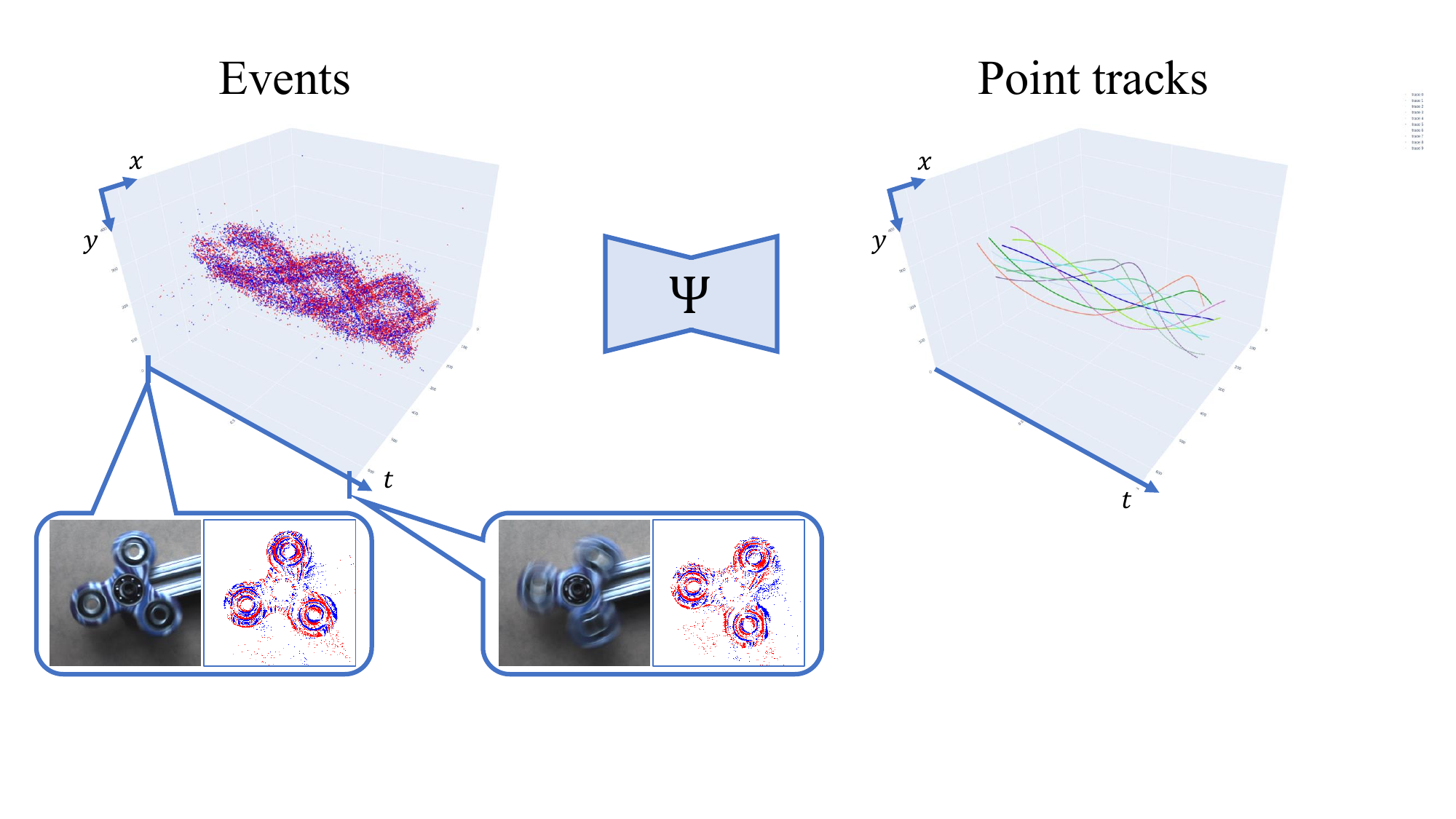}}
\caption{\emph{Event-only Tracking of Any Point}. Our method uses only events to track semi-dense long-range point trajectories, working in conditions where frame-based methods fail.
}
\label{fig:eyecatch}
\vspace{-2ex}
\end{figure}

Estimating point trajectories for arbitrary scene motion presents significant challenges, with recent solutions emerging primarily through deep neural networks trained on synthetic data.
Progress in dense optical flow~\cite{Dosovitskiy15iccv,Teed20eccv} and subsequently point tracking have been driven by supervised learning on rendered datasets, which provide ground truth (GT) scene motion.
Event simulators have enabled a conceptually similar workflow for event-based vision algorithms.
Using the event generation model, events - representing pixel-wise intensity changes - can be synthesized from high-frame-rate videos.
The input video can either be real (temporally upsampled through video interpolation) or synthetic. 
For motion estimation tasks, this approach has been applied to dense optical flow estimation~\cite{Gehrig24pami} and sparse feature tracking~\cite{Messikommer23cvpr}.
However, the synthetic datasets used for training are simplistic warps of 2D objects lacking realism and limiting performance.

\begin{figure}[t]
   \centering
   {\includegraphics[trim={0cm 0cm 0cm 0cm},clip,width=.75\linewidth]{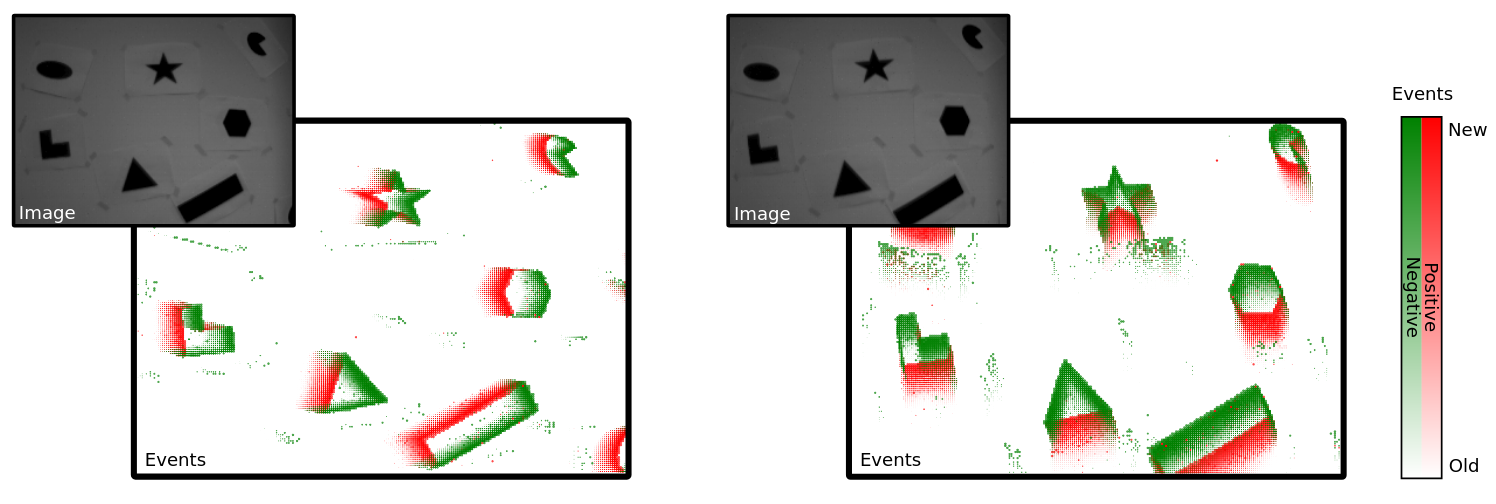}}
\caption{\emph{The motion dependence problem}. Many tracking methods rely on the correspondence of features.
While the appearance of frames (left) is independent of the scene movement, the event camera data depends on the motion direction.
\ifarxiv
Image courtesy of~\cite{Alzugaray22thesis}.
\fi
}
\label{fig:motion_dependence}
\vspace{-2ex}
\end{figure}

The feasibility of training motion-estimation networks using synthetic data has been established for conventional cameras.
Networks demonstrate good generalization, partly by exploiting the correlation of appearance features between timesteps.
With event simulators, similar approaches could be extended to event cameras, though several unique questions must be addressed. 
A key challenge in event-based tracking using synthetic data stems from the immaturity of simulation tools.
While it is possible to combine frame-based physically based rendering (PBR) pipelines with video-to-event conversion tools, this process involves numerous parameters that require tuning to achieve optimal results.
A second challenge with event camera data is its inherent motion dependence.
Consider a simple scenario, illustrated in \cref{fig:motion_dependence}, showing two recordings of an identical scene (e.g. shapes on a wall) with perpendicular camera motions. 
In the first recording, the camera moves horizontally, in the second, vertically.
With a conventional camera, both scenarios would capture nearly identical images (aside from a slight offset).
However, an event camera produces markedly different signals in each case due to its motion-dependent nature.
This poses a unique challenge for algorithms that rely on feature correlation, as the feature extractor must be invariant not only to appearance changes and geometric transformations but \textit{also} to scene motion.

We present, to the best of our knowledge, the first model for event-based tracking of any point (ETAP).
The model tracks several points in parallel, iteratively updating position and appearance features for each point through spatial and temporal attention blocks.%
It processes event stacks (grid representations compatible with convolutional feature encoders), constructed at specified tracking timesteps.

The network is trained on a newly developed synthetic dataset. 
Our data generation pipeline combines Kubric~\cite{Greff22cvpr} and Vid2e~\cite{Gehrig20cvpr}.
Through a systematic evaluation of each design decision (including threshold selection, scene dynamics, and render frame rate) we demonstrate that our {\dname} dataset improves performance by \gred{8\%} (measured by feature age on the Event-aided Direct Sparse Odometry dataset~\cite{Hidalgo22cvpr} (EDS)) compared to the same model trained on the strong baseline approach using pre-rendered Kubric Multi-Object Video (MOVi) - F dataset.

We also introduce a novel contrastive loss that promotes motion-robust feature extraction in our network.
For each training sample, we generate a variant with inverted time and random rotation while preserving appearance.
This transformation maintains the scene structure but inverts motion direction.
We extract spatial feature maps from both representations, interpolate features at tracked points, and reward high cosine similarity between corresponding feature vectors.
Our \lname-loss encourages the generation of motion-invariant correlation features. 

We evaluate two tasks, TAP (Task 1) and additionally on feature tracking (Task 2) for comparison with previous event-based methods.
TAP is evaluated on {\dname}, the Extreme Event Decompression Dataset (E2D2)~\cite{Wang23arxiv_penn} (for which we provide new ground truth), and on custom sequences recorded with a beamsplitter system for fair comparison between event- and frame-based algorithms.
Feature tracking is evaluated on an established benchmark (comprising of EDS and the ``event camera dataset'' (EC)), where \mname~achieves significant improvements over previous event-only methods (\gred{20\%}) and surpassing the best method combining frames and events by \gred{4.1\%}.

Our contributions are summarized as follows:
\begin{enumerate}
    \item The first event-only tracking-any-point (TAP) method, with SOTA results on two tasks (TAP and feature tracking) and extensive evaluation on six datasets (EVIMO2, EDS, EC, E2D2, {\dname}, Aviary).%

    \item A new synthetic event dataset ({\dname}) that enables robust tracking performance, with a thorough empirical evaluation of key design decisions.

    \item For evaluation, we release new ground truth for EVIMO2 and E2D2 sequences, as well as a challenging aviary sequence for qualitative evaluation.

    \item A novel contrastive feature-alignment loss that promotes motion-robust feature extraction from event data.

\end{enumerate}

The experiments show strong cross-dataset generalization to different camera types and resolutions, with outstanding tracking capabilities in a variety of conditions.

\section{Related Work}
\label{sec:related}
\def\figmethodwidth{.35\linewidth}
\begin{table*}[t!]
\centering
\adjustbox{max width=\textwidth}{
\setlength{\tabcolsep}{5pt}
\begin{tabular}{l*{12}{S[table-format=1.3,detect-weight,detect-mode]}}
\toprule
& & & & & & & & \multicolumn{4}{c}{Annotations} \\
\cmidrule{9-12}
{Dataset} & {Source} & {Events} & {\#Samples} & {Resolution} & {\makecell{fps [Hz]}} & {\makecell{sample\\duration [s]}} & {\makecell{IMO}} & {\makecell{optical\\flow}} & {depth} & {\makecell{point\\tracking}} & {segmentations} \\
\midrule %

{TAP-Vid Kubric, MOVI-F}   &  %
{3D PBR}                &  %
{none}                     &  %
{$\approx10000$}           &  %
{$512 \times 512$}         &  %
{12}                       &  %
{2}                        &  %
\cmark                     &  %
\cmark                     &  %
\cmark                     &  %
\cmark                     &  %
\cmark                     \\ %

{BlinkFlow~\cite{Li23iros}}                &  %
{3D PBR}       &  %
{synthetic}             &  %
{3587}                     &  %
{$640 \times 480$}         &  %
{10}                       &  %
{1}                        &  %
\cmark                     &  %
\cmark                     &  %
\cmark                     &  %
\xmark                     &  %
\cmark                     \\ %

{MultiFlow~\cite{Gehrig24pami}}   &  %
{2D warp}                  &  %
{synthetic}                &  %
{12100}                    &  %
{$512 \times 384$}         &  %
{100}                      &  %
{0.5}                        &  %
\cmark                     &  %
\cmark*                    &  %
\xmark                     &  %
\xmark                     &  %
\xmark                     \\ %
{\textbf{\dname~(Ours)}}   &  %
{3D PBR}       &  %
{synthetic}                &  %
{10173}                    &  %
{$512 \times 512$}         &  %
{48}                       &  %
{2}                          &  %
\cmark                     &  %
\cmark                     &  %
\cmark                     &  %
\cmark                     &  %
\cmark                     \\ %
\bottomrule
\end{tabular}
}
\caption{\emph{Dataset comparison}. Overview of publicly available synthetic motion estimation event datasets.
}
\label{tab:dataset-comparison}
\vspace{-2ex}
\end{table*}

\subsubsec{Motion Estimation and Point Tracking.}
Visual motion estimation remains a central theme of general scene understanding, which has, throughout the years, developed into a diverse field of study. 
Early paradigms~\cite{Lucas81ijcai,Baker03part2} focused on estimating the long-term motion of distinctive patches throughout a set of images using an autoregressive template tracking framework. They successively estimated warps from template to target patches in each new image, by minimizing the change in appearance and then updated these templates at each step. 
In the real world, however, image patches often \emph{do change in appearance}~\cite{Matthews04tpami} or distort in complex ways that require the development of complex warping models~\cite{Li13acm}. Moreover, image patches with few gradients often provide insufficient constraints to accurately estimate motion, due to the aperture problem. Increasing the context via variational approaches that optimize a global objective~\cite{Horn81ai} can address this but at the cost of over-smoothing the result. 

Since the advent of deep learning, this context is now captured via deep architectures with large receptive fields~\cite{Ilg17cvpr,Teed20eccv} and regularized by implicit priors learned from data. 
This enabled the tracking of large semantic object bounding boxes ~\cite{Pang21cvpr,Fischer22arxiv}, action bounding boxes~\cite{Cheng22eccv}, or deformable semantic masks~\cite{Pont17arxiv,Kristan21iccv,Wu23iccv}, but these are often constrained to specific object classes and not generally applicable to, for example, object parts or single points. 

Recently, tracking \emph{single points} has gained traction, due to its flexibility in addressing arbitrary structures. For a given point set in a frame, it predicts their corresponding positions in other frames jointly with explicit visibility.
After the early model-based approach Particle Video~\cite{Sand08ijcv} it was re-introduced by ``Particle Video Revisited'' (PIPs)~\cite{Harley22eccv}.
While leveraging many of the early insights for long-term feature tracking such as feature correlation, appearance change modeling, and autoregressive tracking, it did so with modern tools like learning-based feature correlation and a lookup originally designed for optical flow~\cite{Teed20eccv}.

A key driver of this field has been the curation of large-scale synthetic data: The usage of simulated data is scalable, supports dense motion annotations, provides controllable data complexity, and poses fewer problems regarding privacy and licensing.
FlyingChairs~\cite{Dosovitskiy15iccv} and FlyingThings~\cite{Mayer16cvpr} are early datasets widely used for training of optical flow methods, while Kubric provides a flexible dataset generator~\cite{Greff22cvpr} for large-scale training of point tracking.
Follow-up work~\cite{Doersch22neurips,Doersch23iccv} introduced TAP-Vid, a set of synthetic and real-world datasets which form a common benchmark today, along with methods TAP-Net~\cite{Doersch22neurips} and TAPIR~\cite{Doersch23iccv} which innovated on the original design of PIPs. 
Since then PointOdyssey~\cite{Zheng23iccv} appeared, which enhances the realism of the synthetic sequences, and provides additional annotations beyond point tracks.
These benchmarks sparked the development of methods like LocoTrack~\cite{Cho24eccv} and CoTracker~\cite{Karaev24eccv}, which are the state-of-the-art in point tracking.
The work in \cite{Karaev24eccv}, for instance, uses a single model to track several points in parallel leveraging spatial attention between points to model interrelation between them.

\subsubsec{Event Camera-based Motion Estimation.}
Despite these developments, image-based tracking still suffers from fundamental limitations of frame-based sensors, namely a limited framerate, motion blur, and saturation artifacts in challenging lighting conditions, which cause visual aliasing and algorithm degradation. 
Event cameras~\cite{Lichtsteiner08ssc,Posch08iscas} are relatively new vision sensors, which can address these issues with their higher dynamic range, limited motion blur, and ability to capture sparse and asynchronous \emph{changes} in the visual data, also called \emph{events}, in continuous time~\cite{Gallego20pami,Shiba23pami}. 

Similar to image-based tracking, early methods for tracking with events focused on tracking blobs~\cite{Litzenberger06dspws} or simple patterns~\cite{Ni15neco,Lagorce15tnnls}. 
They use iterative closest point (ICP)~\cite{Kueng16iros} or expectation Maximization (EM)~\cite{Zhu17icra} to align small spatio-temporal event volumes or perform multi-hypothesis tracking~\cite{Alzugaray18threedv} that predict the feature motion. 
A main challenge in event camera-based tracking is the dependence of feature appearance on camera and object motion, which limits the use of purely appearance-based trackers.
To address this, appearance refinement~\cite{Alzugaray18threedv,Tedaldi16ebccsp}, auxiliary sensors providing motion-invariant appearance~\cite{Gehrig18eccv,Gehrig19ijcv,Kueng16iros} and data-driven approaches have been explored~\cite{Messikommer23cvpr,Chiberre21cvpr,Manderscheid19cvpr,Liu24arxiv}.
Despite their promise, these methods have their limitations: Refinement and learning-based point trackers still use simple synthetic datasets based on moving 2D planes~\cite{Liu24arxiv,Gehrig24pami,Messikommer23cvpr} which show only a weak transfer to the real world, and thus necessitate self-supervised finetuning~\cite{Messikommer23cvpr,Hamann24eccv}.
On the other hand, methods using frames and events, such as~\cite{Liu24arxiv} specifically combine events and frames in a data-driven approach for point tracking but inherit some of the shortcomings of frames during high-speed motion and in challenging lighting conditions.

In this work, we perform purely event-based tracking and are free of these limitations.
Moreover,
we provide a large-scale, realistic point-tracking dataset for events, summarized in Tab.~\ref{tab:dataset-comparison}.
It enables the learning of powerful priors, together with our novel contrastive feature alignment loss to explicitly enforce motion-independent features across time.

\section{Tracking Any Point With an Event Camera}
\label{sec:method}

\begin{figure*}[t]
	\centering  %
    {\includegraphics[trim={0cm 5.2cm 0.2cm 4.1cm},clip, width=\linewidth]{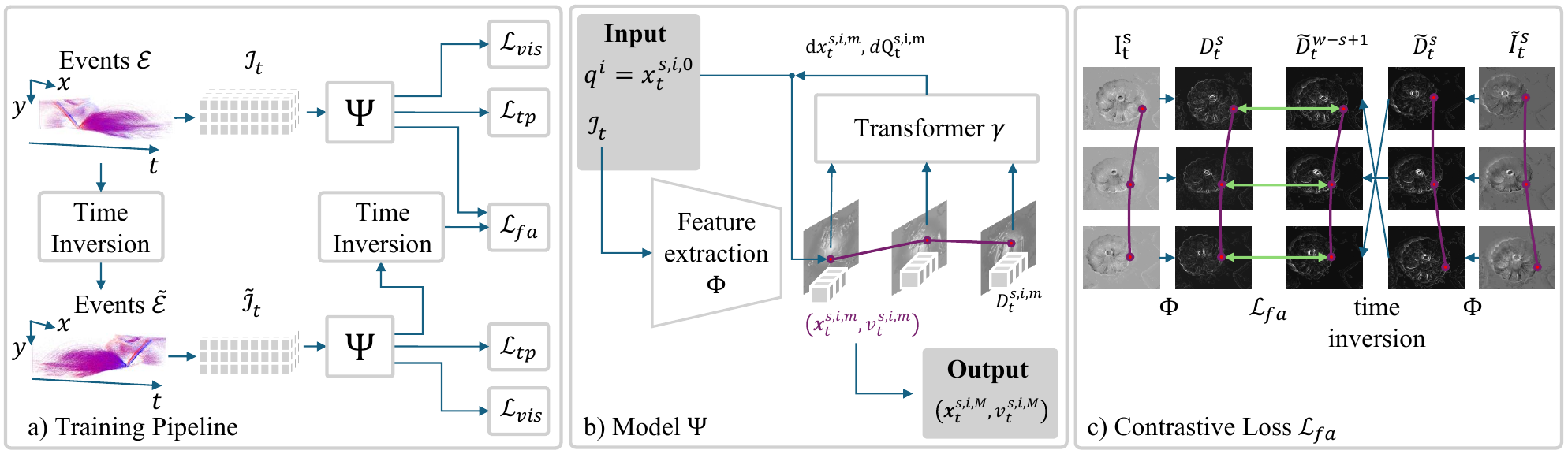}}
	\caption{(a) During training each sample has a time inverted duplicate, model $\Psi$ outputs tracks, visibility flags, and descriptors used to calculate the 3 loss terms.
    (b) The model $\Psi$ takes query points and event stacks as input and extracts spatial feature maps at each timestep. tokens built per point and timestep are iteratively updated.
    (c) A visualization of the FA-loss. Descriptors of the inverted second sample are time-reversed to receive matching pairs for the FA-loss}
\label{fig:method}
\vspace{-2ex}
\end{figure*}

\subsubsec{Problem Formulation.}
Let us formalize the TAP task with event cameras.
These sensors measure so-called \emph{events}, \emph{i.e.} per-pixel brightness changes $e_k \doteq (\bx_k, \tau_k, \pol_k)$ where $\bx_k = (x_k,y_k)^\top$ is the pixel the event is produced, 
$\tau_k\in\Real$ is its timestamp with \si{\micro\second} resolution and $p_k\in\{-1, 1\}$ is its polarity (the sign of the brightness change).
Each event $e_k$ is triggered when the logarithmic brightness at pixel $\bx_k$ exceeds a threshold $C$, called contrast sensitivity. 
Over a time interval $\mathcal{T}=(\tau_s, \tau_e)$ the event camera thus outputs asynchronous events $E=\{e_k\}$ at different pixels. 

Next, let $P(\tau)=\{(\bx^i(\tau), Q^i(\tau), v^i(\tau))\}_{i=1}^{N_p}$ be a set of $N_p$ points moving over time $\tau\in\mathcal{T}$, where $\bx^i(\tau)\in\Real^2$ is the pixel position of point $i$, $Q^i(\tau)\in\Real^d$ is its descriptor and $v^i(\tau)\in \{0, 1\}$ is its visibility.
The descriptors $Q^i$ are 1D feature vectors used to estimate visual similarity with per-timestamp feature maps.
Separate descriptors of the same point at different times, allow modeling appearance changes.
Note that points may be initialized asynchronously, and thus the cardinality of $P(\tau)$ may not remain constant over time.

Following the formalism in \cite{Gehrig24pami} we focus on the point positions at discrete time instances $\tau_t$ with $t=0,1,...,T$ and define their position at these instances by $P_t \doteq P(\tau_t)$.
Similarly, we select windows of events 
\begin{equation}
E_t = \{e_k \, | \, \tau_k\in(\tau_t-\Delta \tau _t,\tau_t)\}\subset E
\end{equation}
that are temporally aligned with $\tau_t$, %
where $\Delta \tau_t$ is the time span of $E_t$, which contains a constant number of events $N_e$.
We consider a sliding window of such point observations $\cP_{t} \equiv P_{t-w+1:t}%
$, with window size $w=8$,as well as the sequence $\cE_{t} \equiv E_{t-w+1:t}$ defined similarly. 
We formalize TAP as finding the function $\Psi$ that estimates the tracks $\cP_{t}$ from events $\cE_{t}$ and past tracks $\cP_{t-T_s}$
\begin{equation}
    \label{eq:tracker}
    \cP_{t} = \Psi(\cP_{t-T_s},\, \cE_{t}).
\end{equation}
where $T_s=4$ is the stride.

\subsubsec{Feature Representations.}
In practice, each event window $E_{t}$ is replaced by event representations $I_t = \mathcal{F}(E_t) \in \Real^{H\times W \times B}$ \cite{Nam22cvpr} where $H$ and $W$ are the sensor's height and width, and $B=10$ is the number of time bins.
\cite{Gehrig19iccv,Deng22cvpr,Zubic23iccv}. 

We extract multi-scale $d$-dimensional features $D^\lambda_t\in\Real^{\frac{H}{k2^{\lambda-1}}\times \frac{W}{k2^{\lambda-1}}\times d}$ from tensors $I_t$ using an encoder $\phi(I_t)$, with subsequent average pooling.
$\lambda=1,...,S$  (with $S=4$) is the scale, and $k=4$ an overall reduction in the resolution.

\subsubsec{Initialization.}
We manually provide query points $q^i = \bx^i_{t_i}$ at time indices $t_i$, and before the subsequently explained transformer-based refinement, broadcast the points to all timesteps $\bx^i_{t}=\bx_{t}$ of the sliding window, where a point is initialized.
Similarly, the descriptors $Q^i_t$ are initialized via the broadcast $Q^i_t = Q^i_{t_i}$, with $Q^i_{t_i} = \text{BilinearInterp}(D^\lambda_{t_i}, \bx^i_{t_i})$, where $\text{BilinearInterp}(\cdot, \cdot)$ samples the feature map $D^\lambda_{t_i}$ at continuous coordinates $\bx^i_{t_i}$ using bilinear interpolation.

\subsubsec{Tracker.}
We implement the tracker \eqref{eq:tracker} $\Psi$ following \cite{Karaev24eccv}.
For simplicity, we omit the global timestep $t$ and regard only point positions within one sliding window $P_s^i \doteq P_{t-w+s}^i$ with relative window index $s=1,\ldots,w$, and $\mathcal{P}\doteq P^i_{1:w}$.
We denote $\cD \doteq D^\lambda_{t-w+s:t}$ as the according feature maps defined in the same interval as the point tracks $\cP$.

Specifically, the tracker iteratively refines pixel positions $\bx^i_s$ and descriptors $\tilde{Q}^i_s$ of the $i^{th}$ point via  
\begin{align}
    ( \text{d}\tilde{\bx}^{i,m}_s, \text{d}\tilde{Q}^{i,m}_s ) & = \gamma(\cP^{m}, \cD)\\
    \tilde{\bx}^{i,m+1}_s &= \tilde{\bx}^{i,m}_s + 
    \text{d}\tilde{\bx}^{i,m}_s\\
   \tilde{Q}^{i,m+1}_s &= \tilde{Q}^{i,m}_s + 
    \text{d}\tilde{Q}^{i,m}_s
\end{align}
Note that $\tilde{Q}^{i,m}_s$ and $\tilde{\bx}^{i,m}_s$ denote the descriptor and position at iteration $m$, and $\cP^{m}$ are the tracks with updated descriptor and point position.
The update step is iterated $M$ times, to obtain the final position estimates $\tilde{\bx}^{i,M}_s$.
After the last iteration, visibilities are computed with a simple linear layer via $v^i_s=\sigma(\Theta\tilde{Q}^{i,M}_s)$.

We implement $\gamma$ as a transformer that operates on tokens $\mathcal{O}^{i,m}_s$ indexed by relative time $s$ and point index $i$ via alternating intra-point attention (across index $i$), and temporal attention (across index $s$).
At each iteration $m$ we compute these tokens as the following concatenation:%
\begin{align}
\nonumber\mathcal{O}^{i,m}_s=\left(\eta(\Delta\bx_s^{i,m}), Q_s^{i,m}, C_s^{i,m}, v_s^{i}\right)
    \nonumber+ \eta'(\bx_0^{i,m})+\eta'(t)
\end{align}
 with $\Delta\bx_s^{i,m}= \bx_s^{i,m}-\bx_1^{i,m}$, positional encodings $\eta,\eta'$, and spatial correlation features $C_s^{i,m}$, which are discussed next.

\subsubsec{Correlation Features.}
As part of the input tokens to the transformer, we provide information on the similarity of descriptors $\tilde{Q}^{i, m}_s$ to points of their surroundings.
The correlation features within a patch $B_\Delta=\{\delta \in \mathbb{Z}^2 \vert \, \|\delta\|_\infty \leq \Delta\}$ are calculated via the inner products 
\begin{equation}
    C_s^{i, m} = \oplus_{\lambda=1}^S \oplus_{\delta \in B_\Delta} \left\langle 
\tilde{Q}_s^{i, m}, D(\tilde{\bx}_{s}^{i, m}/k\lambda+\delta) \right\rangle
\end{equation}
Here we are concatenating correlations within the patch $B_\Delta$ of size $\vert B_\Delta \vert=(2\Delta+1)^2=49$ across four scales, resulting in a feature dimension of $196$.
At each iteration $m$ of the transformer refinement, we use the updated point locations $\tilde{\bx}_s^{i,m}$ to compute these features.

\subsubsec{Motion Robust Event Features.}
Event camera data is inherently motion-dependent, unlike conventional cameras where the same scene produces the same signal regardless of motion (\cref{fig:motion_dependence}).
Based on the linearized event generation model (LEGM), we can show that under time inversion, the events $E_t$ and events of the inverted scene $\tilde{E}_t$ are not the same (see Supplementary for mathematical derivation), and consequently, the corresponding descriptors $D_t^s$ and $\tilde{D}_t^{w-s+1}$ are different (note $w-s+1$ is the inverted index).
We explicitly enforce motion consistency by maximizing the similarity of descriptors ${d}_t^{s,i}\doteq D_t^{s} (\bx_t^{s,i})$ and 
$\tilde{d}_t^{s,i}\doteq \tilde{D}_t^{w-s+1} (\bx_t^{w-s+1,i})$ 
sampled at track positions $\bx_k^{s,i}$ and $\bx_k^{{w-s+1,i}}$, under different motions, and incorporate this into the loss function: 
\begin{equation}
\label{eq:fa_loss}
\textstyle
    \mathcal{L}_\text{fa} = \sum_t  \frac{1}{|\cP_t|} 
    \sum_{i,s}\left(1-\bigl\langle \operatorname{u}{\bigl( {d}_t^{s,i} \bigr)}, 
    \operatorname{u}{\bigl( \tilde {d}_t^{s,i} \bigr)} \bigr\rangle \right)^2
\end{equation}
where $\operatorname{u}(\mathbf{a}) = \mathbf{a} / \| \mathbf{a} \|$ unitizes a vector, 
and $|\cP_t|$ counts the number of points within the time window. 
To further enhance the diversity of motion, we randomly rotate the events $\tilde{E}$ by an angle $\theta\in \{0, 90^\circ,180^\circ,270^\circ\}$.

We supplement this loss with the track prediction error $\mathcal{L}_\text{tp}$, which penalizes the absolute difference between predicted and GT tracks at each refinement step $m$ weighted by $0.8^{M-m}$.
We also include the visibility loss $\mathcal{L}_\text{vis}$, which is the cross-entropy on predicted visibility flags~\cite{Karaev24eccv}.
Our total loss is calculated as $\mathcal{L} = 0.1\mathcal{L}_\text{tp} + \mathcal{L}_\text{vis} + 0.1\mathcal{L}_\text{fa}$.

\section{Dataset Generation}
\label{sec:data}

\def\figWidth{0.38\linewidth}
\begin{figure}[t]
    \centering
    {\footnotesize
    \setlength{\tabcolsep}{1pt}
    \begin{tabular}{
    >{\centering\arraybackslash}m{0.26cm} 
    >{\centering\arraybackslash}m{\figWidth} 
    >{\centering\arraybackslash}m{\figWidth}
    }
        \rotatebox{90}{\makecell{GT}}
        &\gframe{\includegraphics[clip,trim={0 0 0 0},width=\linewidth]{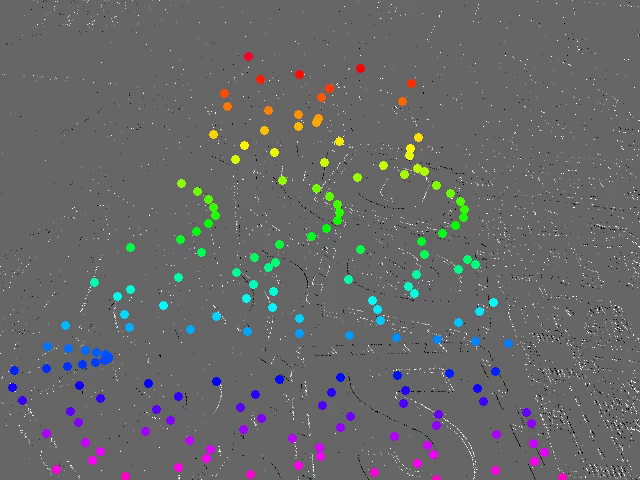}}
        &\gframe{\includegraphics[clip,trim={0 0 0 0},width=\linewidth]{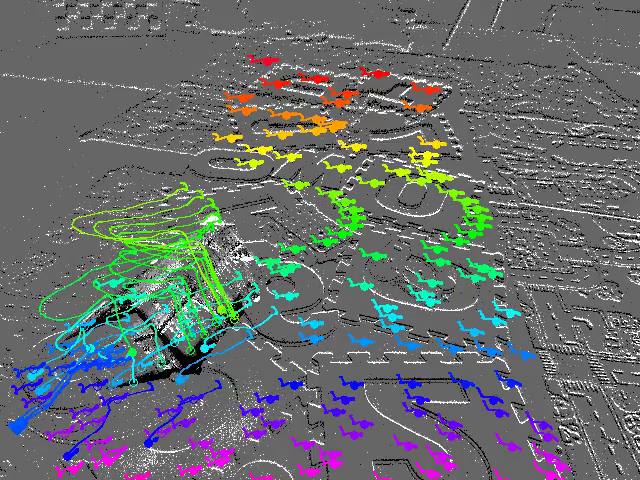}} \\[1ex]
        \rotatebox{90}{\makecell{Ours}}
        &\gframe{\includegraphics[clip,trim={0 0 0 0},width=\linewidth]{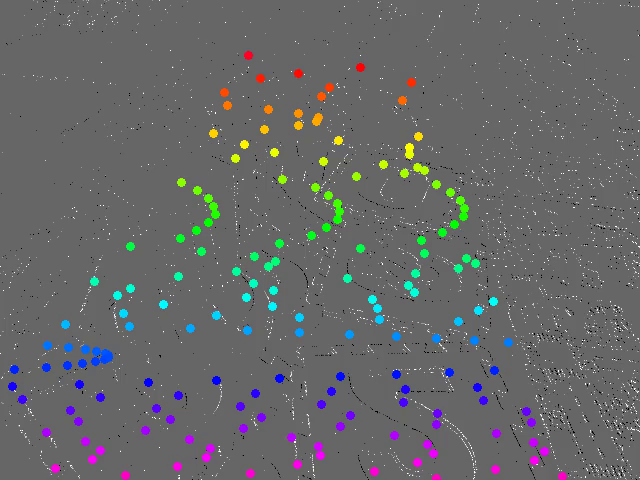}}
        &\gframe{\includegraphics[clip,trim={0 0 0 0},width=\linewidth]{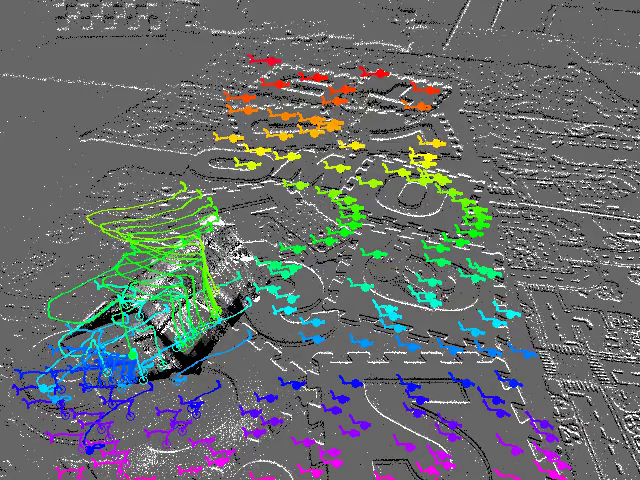}} \\[-0.5ex]
        & $t = 0s$
        & $t = 1.5s$
    \end{tabular}
    }
    \vspace{-2ex}
\caption{\emph{Task 1 - TAP on EVIMO2}.
Visualization of track predictions from first to last timestamp.
}
\label{fig:evimo_pred_reduced}
\end{figure}
\begin{figure}[t]
   \centering
   {\includegraphics[trim={3cm 4.3cm 2.5cm 4.5cm},clip,width=.9\linewidth]{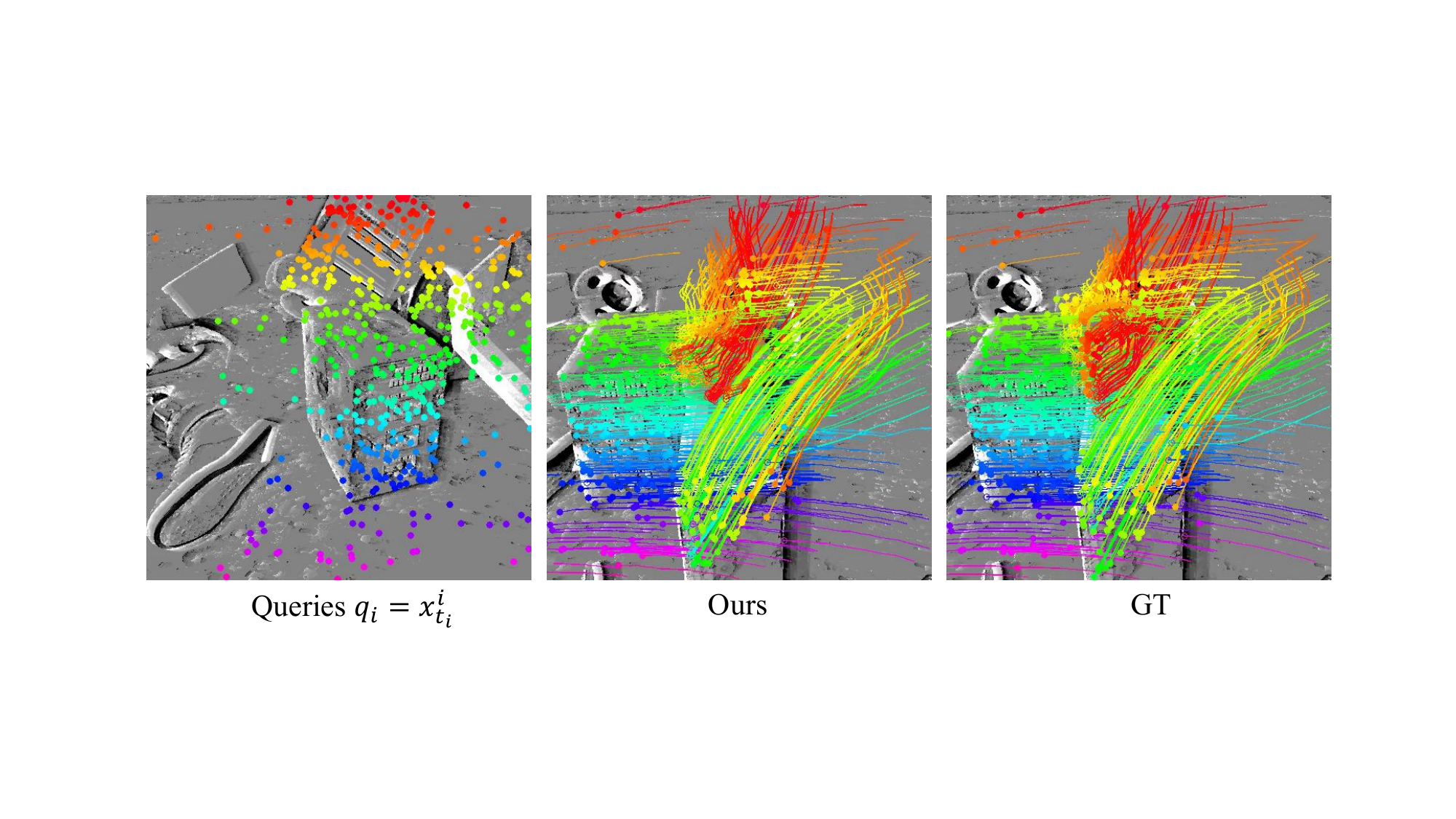}}
   \vspace{-2ex}
\caption{\emph{Task 1 - TAP on {\dname}}.
Semi-dense tracks are predicted for 2s-samples.
}
\label{fig:event_kubric_pred}
\vspace{-1ex}
\end{figure}

\begin{table}[t!]
\centering
\adjustbox{max width=0.9\linewidth}{
\setlength{\tabcolsep}{4pt}
\begin{tabular}{lc*{3}{S[table-format=1.3,detect-weight,detect-mode]}}
\toprule
& & \multicolumn{3}{c}{Metrics} \\
\cmidrule(lr){3-5}
{Method} & {Input} & {AJ $\uparrow$} & {$\delta^x_{avg} \uparrow$} & {OA $\uparrow$} \\
\midrule
\multicolumn{5}{c}{\textbf{EVIMO2}} \\
\midrule
{E2Vid~\cite{Rebecq19cvpr} + CoTracker~\cite{Karaev24eccv}}  & Events & {0.531} & {0.663} & {0.861} \\
\textbf{\mname~w\textbackslash o FA-loss~(Ours)} & Events & \unum{1.3}{0.655} & \unum{1.3}{0.787} & \unum{1.3}{0.884} \\
\textbf{\mname~(Ours)}                           & Events & \bnum{0.661} & \bnum{0.789} & \bnum{0.895} \\
\midrule
\multicolumn{5}{c}{\textbf{\dname~(synthetic)}} \\
\midrule

{E2Vid~\cite{Rebecq19cvpr} + CoTracker~\cite{Karaev24eccv}}  & Events & {0.236} & {0.331} & {0.815} \\
\textbf{\mname~w\textbackslash o FA-loss~(Ours)} & Events & \bnum{0.556} & \bnum{0.677} & \bnum{0.894} \\
\textbf{\mname~(Ours)}                           & Events & \unum{1.3}{0.546} & \unum{1.3}{0.675} & \unum{1.3}{0.890} \\
\bottomrule
\end{tabular}
}
\vspace{-1ex}
\caption{\emph{Task 1 - TAP evaluation on {\dname} and EVIMO}.}
\label{tab:results_event_kubric}
\vspace{-3ex}
\end{table}

Our model training relies on a new synthetic dataset created through a three-step process:
First, we render short video clips using Kubric~\cite{Greff22cvpr}, then adaptively upsample them using FILM~\cite{Reda22eccv}, and finally convert the resulting high-frame-rate video to events using ESIM~\cite{Rebecq18corl}.
For each sample, we generated 2048 point tracks derived from Kubric's ground truth data.
The complete dataset comprises 10,173 samples, split randomly into 80:15:5 ratios for training, validation, and testing.
Representative samples from the training set are provided in the Supplementary material.

\subsubsec{Physics-based rendering.}
Using Kubric~\cite{Greff22cvpr} we render 2-s videos at 48 FPS resulting in 96 frames with 512 $\times$ 512-px resolution.
We opt for a higher FPS than the available Kubric datasets, and disable motion blur to reduce the error introduced through upsampling and event generation.
Scenes contain approximately 20 3D rigid objects under gravity simulated with the BULLET~\cite{Coumans15sigg} physics engine and ray-tracing with Blender~\cite{Blender18blender}.
We generate 60\% of samples with linear camera movement and 40\% with panning movements as used in~\cite{Doersch23iccv}, mimicking natural camera movements found in many real datasets.

\subsubsec{Synthetic Event Generation.}
Due to the computational cost of ray-traced rendering and Kubric's fixed frame rate constraint, we adopt the VID2E~\cite{Gehrig20cvpr} workflow, employing adaptive neural frame interpolation, such that the maximum optical flow between consecutive upsampled frames is at most one pixel, following~\cite{Rebecq18corl}.
After upsampling, we generate events using random contrast sensitivities $C\sim \mathcal{U}(0.16, 0.34)$ as in~\cite{Klenk24threedv}.

\subsubsec{Point Track Generation.}
Since Kubric doesn't directly provide point tracks, we compute them from the depth, segmentation, and surface normal GT.
Following~\cite{Doersch22neurips}, we randomly sample 2048 GT tracks, ensuring adequate object coverage (compared to a high background pixels portion).

\section{Experiments}
\label{sec:exp}
After showing implementation details in \cref{sec:exp:impl_details} explains, we validate our method on two tasks: TAP (\cref{sec:exp:tap}) and feature tracking (\cref{sec:exp:feat_trk}).
The main technical difference of TAP is the explicit prediction of visibility flags.
Additional evaluation of feature tracking allows for a thorough comparison with state-of-the-art event-based methods on established benchmarks.
Lastly, we present ablations in \cref{sec:exp:sensitivity}.

\subsection{Implementation Details}
\label{sec:exp:impl_details}
Our model was trained on 4 NVIDIA A100 80GB GPUs with a batch size of 2, resolution of 512$\times$512, and the AdamW optimizer~\cite{Loshchilov17arxiv} with a learning rate of $5\cdot 10^{-4}$ and weight decay $10^{-4}$.
Each sample comprises 256 trajectories with length 24.
For the first $10^5$ steps, we optimize only the track prediction and visibility loss and then add $\mathcal{L}_{fa}$ for $1.2 \cdot 10^5$ steps.
For training, on {\dname} we use $N_e = 4 \cdot 10^5$ events (for more info see Suppl.Mat.).

The event stacks undergo std-mean normalization, computed across batch and time dimensions but independently for each channel to accommodate the varying event counts.
We apply Gaussian noise augmentation with $\sigma = 0.1$ (event counts) for the first channel, scaling according to the event count for the remaining channels.

\def\figWidth{0.23\linewidth}
\begin{figure}[t]
	\centering
    {\footnotesize
    \setlength{\tabcolsep}{1pt}
	\begin{tabular}{
	>{\centering\arraybackslash}m{0.26cm} 
	>{\centering\arraybackslash}m{\figWidth} 
	>{\centering\arraybackslash}m{\figWidth} 
	>{\centering\arraybackslash}m{\figWidth} 
	>{\centering\arraybackslash}m{\figWidth} 
    }

        \rotatebox{90}{\makecell{GT}}
		&\gframe{\includegraphics[clip,trim={9cm 5.1cm 6cm 4.5cm},width=\linewidth]{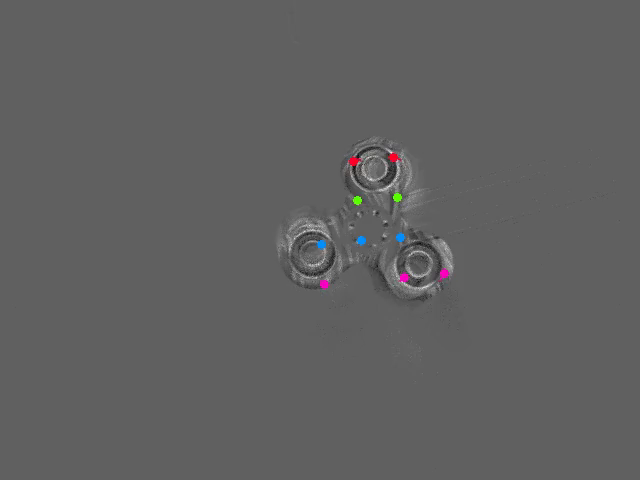}}
        &\gframe{\includegraphics[clip,trim={9cm 5.1cm 6cm 4.5cm},width=\linewidth]{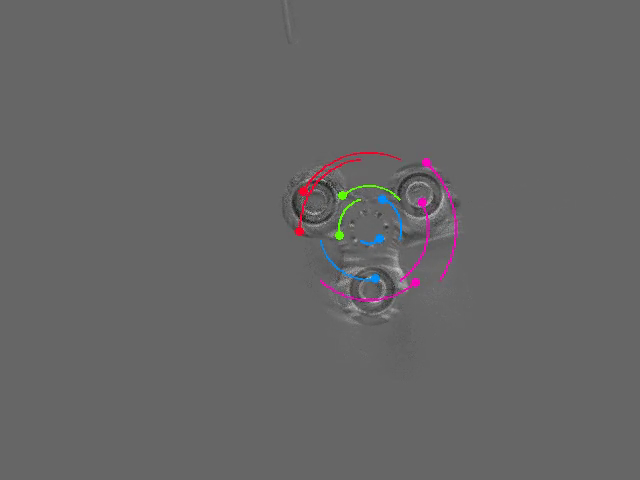}}
		&\gframe{\includegraphics[clip,trim={9cm 5.1cm 6cm 4.5cm},width=\linewidth]{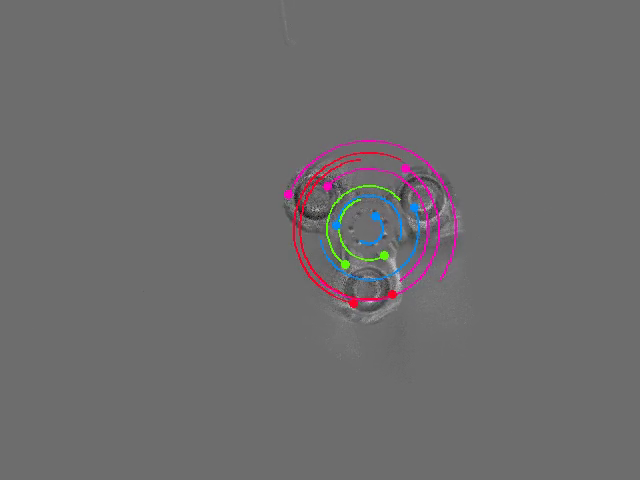}}
        &\gframe{\includegraphics[clip,trim={9cm 5.1cm 6cm 4.5cm},width=\linewidth]{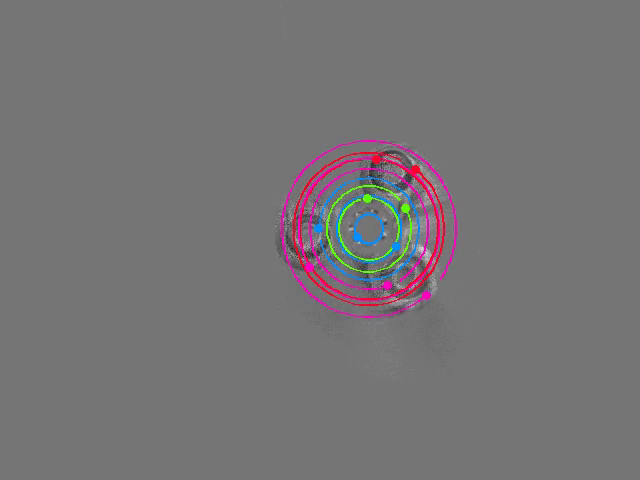}} \\[1ex]

        \rotatebox{90}{\makecell{Ours (E)}}
		&\gframe{\includegraphics[clip,trim={9cm 5.1cm 6cm 4.5cm},width=\linewidth]{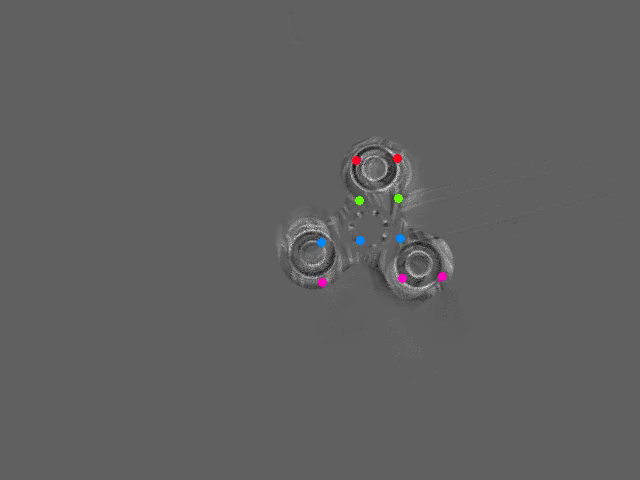}}
        &\gframe{\includegraphics[clip,trim={9cm 5.1cm 6cm 4.5cm},width=\linewidth]{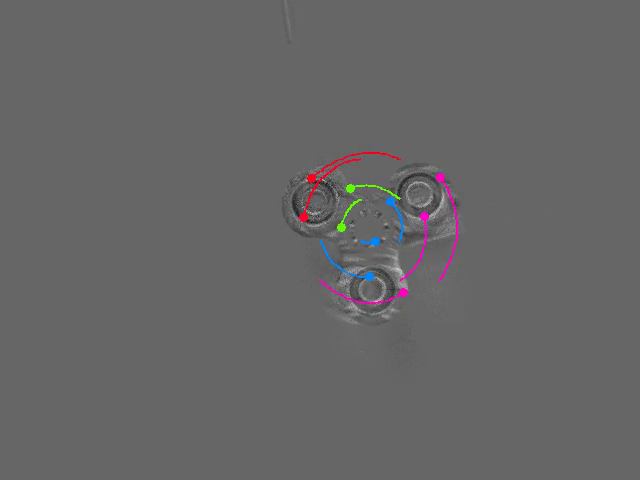}}
		&\gframe{\includegraphics[clip,trim={9cm 5.1cm 6cm 4.5cm},width=\linewidth]{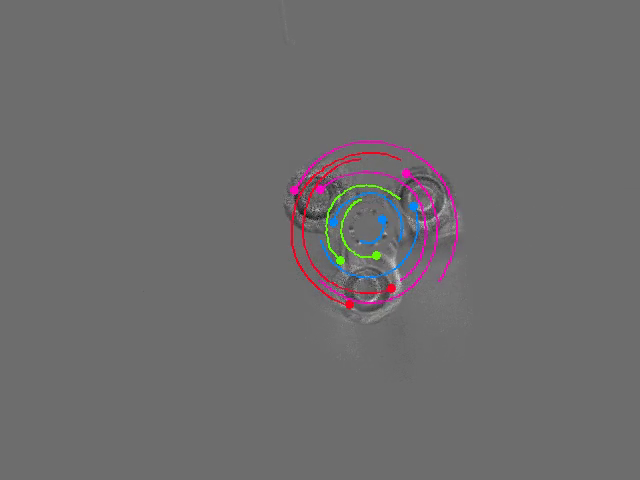}}
        &\gframe{\includegraphics[clip,trim={9cm 5.1cm 6cm 4.5cm},width=\linewidth]{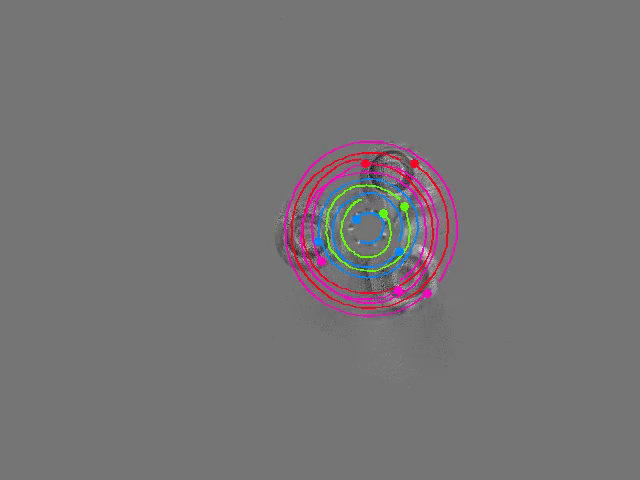}} \\

        \rotatebox{90}{\makecell{E2Vid+CoTr. (E)}}
		&\gframe{\includegraphics[clip,trim={9cm 5.1cm 6cm 4.5cm},width=\linewidth]{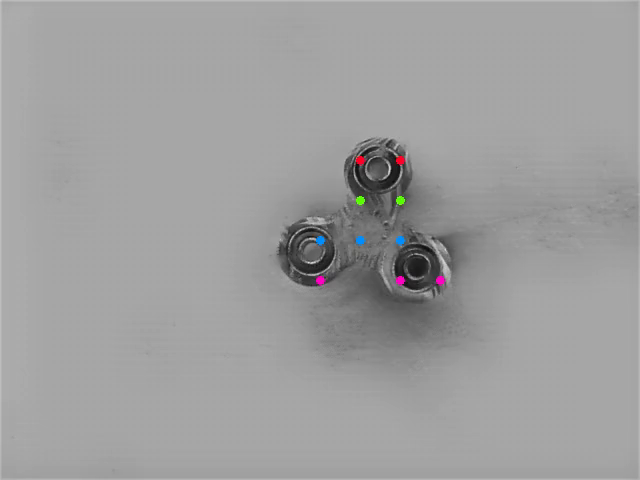}}
        &\gframe{\includegraphics[clip,trim={9cm 5.1cm 6cm 4.5cm},width=\linewidth]{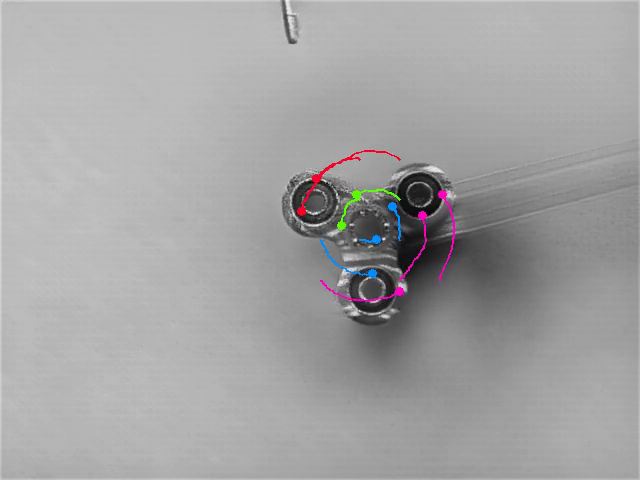}}
		&\gframe{\includegraphics[clip,trim={9cm 5.1cm 6cm 4.5cm},width=\linewidth]{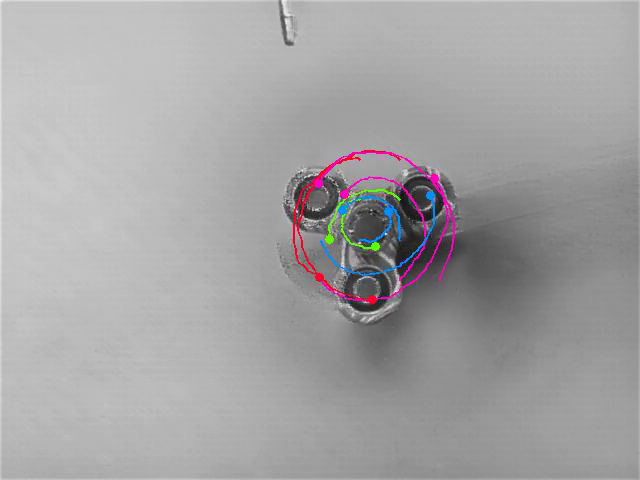}}
        &\gframe{\includegraphics[clip,trim={9cm 5.1cm 6cm 4.5cm},width=\linewidth]{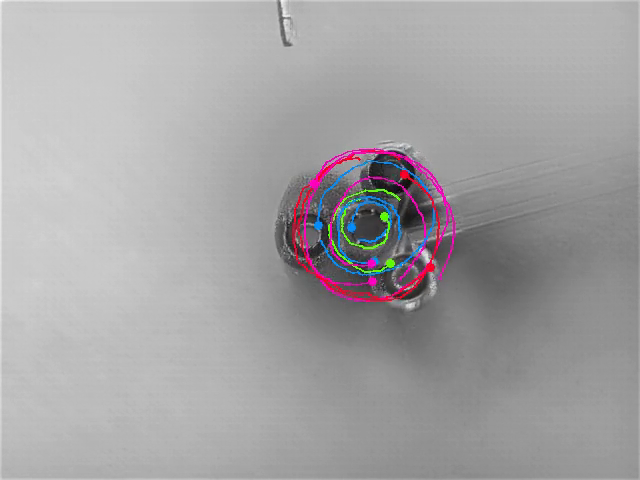}} \\

        \rotatebox{90}{\makecell{CoTracker (F)}}
		&\gframe{\includegraphics[clip,trim={9cm 5.1cm 6cm 4.5cm},width=\linewidth]{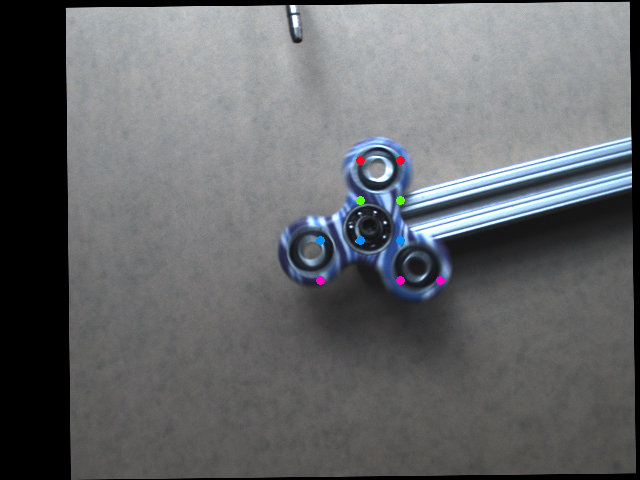}}
        &\gframe{\includegraphics[clip,trim={9cm 5.1cm 6cm 4.5cm},width=\linewidth]{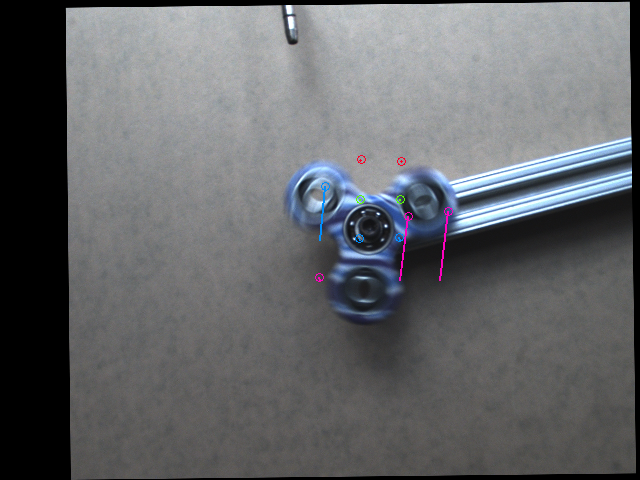}}
		&\gframe{\includegraphics[clip,trim={9cm 5.1cm 6cm 4.5cm},width=\linewidth]{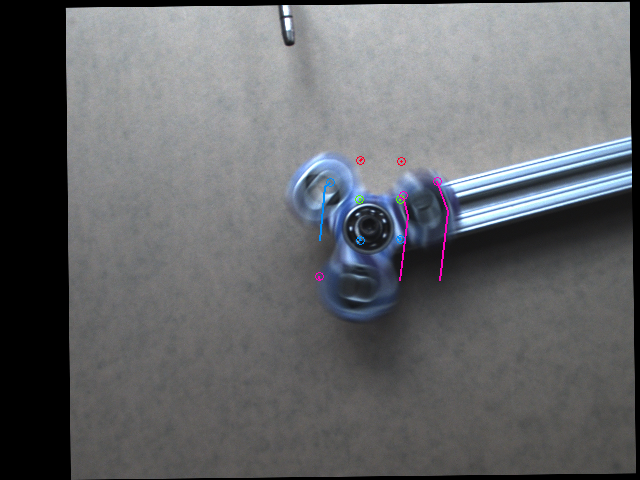}}
        &\gframe{\includegraphics[clip,trim={9cm 5.1cm 6cm 4.5cm},width=\linewidth]{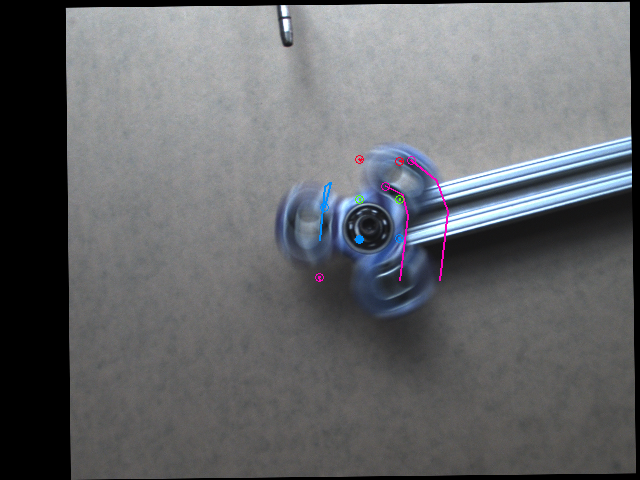}} \\
        
		& \textbf{(a)} $t_0$ (queries)
		& \textbf{(b)} $t_1$
		& \textbf{(c)} $t_2$
		& \textbf{(d)} $t_3$
	\end{tabular}
	}
    \vspace{-1ex}
	\caption{\emph{Task 1 - TAP on E2D2.} Shown are four timesteps of each sequence. At the beginning ($t_0$) the model is queried with the marked points. Input modality: F - frames, E - Events}
	\label{fig:exp:e2d2}
    \vspace{-1ex}
\end{figure}

\begin{table}[t!]
\adjustbox{max width=\linewidth}{
\setlength{\tabcolsep}{4pt}
\begin{tabular}{lc*{3}{S[table-format=1.3,detect-weight,detect-mode]}}
\toprule
& & \multicolumn{3}{c}{Metrics} \\
\cmidrule(lr){3-5}
{Method} & {Input} & {AJ $\uparrow$} & {$\delta^x_{avg} \uparrow$} & {OA $\uparrow$} \\
\midrule
{CoTracker~\cite{Karaev24eccv}} & {F} & {0.007} & {0.117} & {0.1} \\
{E2Vid~\cite{Rebecq19cvpr} + CoTracker~\cite{Karaev24eccv}} & {E} & {0.183} & {0.262} & \bnum{0.938} \\
\textbf{\mname~w\textbackslash o FA-loss~(Ours)} & {E} & \unum{1.3}{0.268} & \unum{1.3}{0.406} & \unum{1.3}{0.826} \\
\textbf{\mname~(Ours)} & {E} & \bnum{0.308} & \bnum{0.466} & {0.769} \\
\bottomrule
\end{tabular}
}
\vspace{-1ex}
\caption{\emph{Task 1 - TAP on fidget spinner (E2D2 dataset).}}
\label{tab:e2d2_fidget_results}
\vspace{-2ex}
\end{table}

\subsection{Task 1: TAP}
\label{sec:exp:tap}

\subsubsec{Results on EVIMO2.}
We evaluate point tracking on real event data using EVIMO2~\cite{Burner22evimo2}, creating new ground truth tracks from its motion capture data.
Our approach mirrors the EVIMO2 Continuous Flow Dataset~\cite{Hamann24eccv} methodology, differing only in our generation of long-term tracks with occlusion flags.
Tests use Samsung event camera data (640 × 480px) featuring independently moving rigid objects in both dynamic scenarios and static conditions where event-based methods typically struggle due to their motion dependence.
\Cref{fig:evimo_pred_reduced} and \cref{tab:results_event_kubric} show the strong performance of our method predicting long tracks and occlusions.

\subsubsec{Results on {\dname}.}
We evaluate performance on our {\dname} test split (501 $\times$ 2s samples), comparing against CoTracker~\cite{Karaev24eccv} operating on E2VID~\cite{Rebecq19cvpr} images.
We select 24 evenly distributed tracking timestamps within each sample and assess performance using standard TAP-metrics~\cite{Doersch22neurips}:
Average-pixel-within-threshold $\delta^x_\text{avg}$ measures the fractions of visible points within a threshold at several levels (1, 2, 4, 8, 16px in 512 x 512 resolution), occlusion accuracy (OA) is the fraction of correct visibility corrections $v^i_t$ and average Jaccard (AJ),  combines both, where a point is correctly predicted when it is within the threshold and has correct visibility prediction.
Quantitative results on \cref{tab:results_event_kubric} demonstrate that our model learns stable track and occlusion prediction, surpassing the E2Vid+CoTracker baseline by \gred{136}\%.
\cref{fig:event_kubric_pred} shows an example prediction.

\subsubsec{Results on E2D2.}
To compare the limitations of frame and event-based algorithms, we design an experiment using the recent Extreme Event Decompression Dataset~\cite{Wang23arxiv_penn} (E2D2), which provides synchronized frames and events from a beamsplitter system.
We utilize this setup for fair cross-modality comparison, specifically picking a scene of a rotating fidget spinner (\cref{fig:exp:e2d2}, where angular velocity increases rapidly over 0.5 seconds, progressively challenging tracking performance.
The scene, recorded under low light conditions, limits the frame-based camera to 10 Hz due to exposure constraints.
Similarly, the event camera data exhibits significant noise and shadow-induced artifacts.
We generate 330 Hz GT tracks for query points on the fidget spinner using angular velocity estimates (see Suppl.Mat.).
We compare against two methods, one is CoTracker~\cite{Karaev24eccv} run on the RGB frames and a second time on E2VID images reconstructed at the GT timestamps.
The results show that the frame-based tracking method fails due to two factors: aliasing from the wheel's rotation combined with low frame rate, and severe motion blur from extended exposure times.
In contrast, the events capture information at sufficient temporal resolution.
Quantitative comparison in \cref{tab:e2d2_fidget_results} shows the comparison of the two event-based methods shows that our~\mname~produces significantly more accurate tracks, surpassing the event-based baseline by \gred{68}\% AJ.

\def\figmethodwidth{.5\linewidth}
\begin{figure}[t]
   \centering
   {\includegraphics[trim={8cm 2cm 12.5cm 0.5cm},clip,width=.8\linewidth]{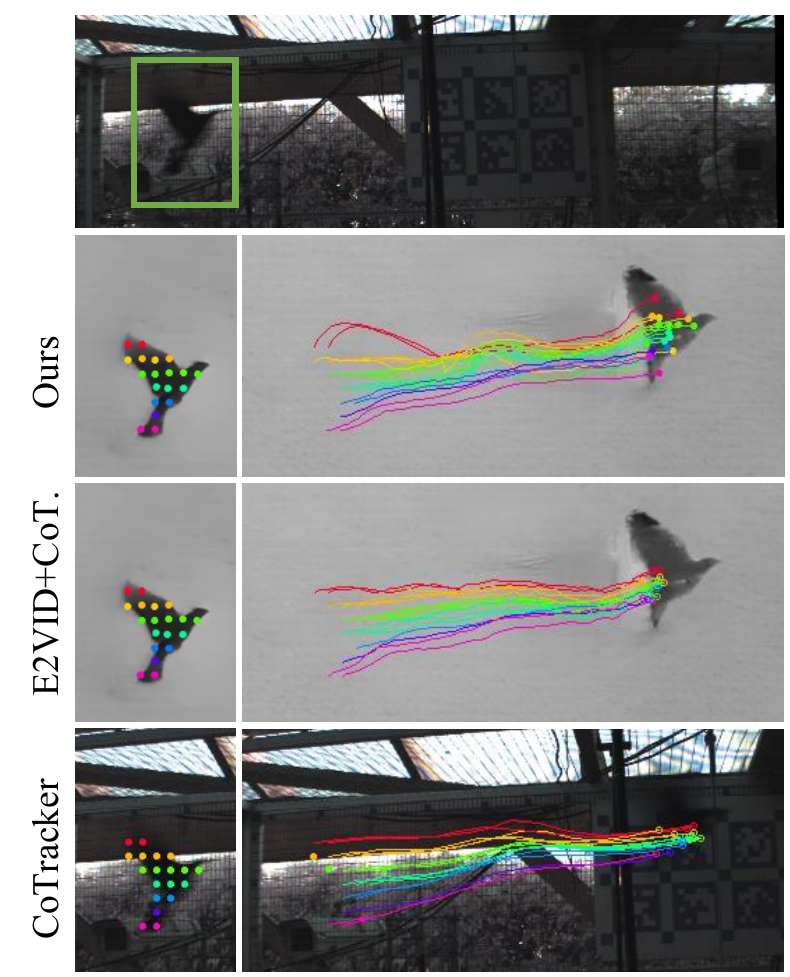}}
\caption{\emph{Task 1 - TAP qualitative result}.
Tracking in a very demanding scenario: a small, low-textured, fast object with high deformation and an HDR background.
}
\vspace{-1ex}
\label{fig:qualitative_results_descriptive}
\vspace{-2ex}
\end{figure}

\subsubsec{Qualitative Analysis.}
We perform qualitative analysis on recordings from E2D2 and additional recordings that were done in a similar manner using a beamsplitter system.
\cref{fig:qualitative_results_descriptive} shows one example.
The sequence of a bird, with RGB frames at 66 Hz is very challenging with a small, fast-moving, highly deforming target with little structure and an HDR background.
The comparison with CoTracker~\cite{Karaev24eccv} a state-of-the-art frame-based model, shows that our model performs better in tracking the target and capturing details.
It also captures more details like the wing flaps, showing the potential of event-based TAP and the improved performance against an event-based E2VID+CoTracker baseline.

\subsection{Task 2: Feature Tracking}
\label{sec:exp:feat_trk}

\subsubsec{Results on EDS \& EC.}
We evaluate our method on the EC~\cite{Mueggler17ijrr} EDS ~\cite{Hidalgo22cvpr} datasets, following standard feature tracking protocols.
These datasets provide synchronized events and frames at resolutions of 240$\times$180 and 640$\times$480 px respectively.
Performance is measured using \textit{feature age} (FA) and \textit{expected feature age}, which quantify the duration until a track deviates beyond a threshold distance from the GT.
For detailed descriptions of the evaluation protocol and metrics, we refer to ~\cite{Messikommer23cvpr}.
We evaluate our tracker against two categories of methods: those using only events and those using events and frames for enhanced information.
Event and frame-based methods comprise ICP~\cite{Kueng16iros}, ``Event-based Kanade-Lucas-Tomasi'' (EKLT)~\cite{Gehrig19ijcv}, which employs template patches extracted from grayscale frames with subsequent event-based tracking, ``Data-driven feature tracking for Event Cameras'' (DDFT)~\cite{Messikommer23cvpr}, a recent data-driven approach using similar principles, and ``Frame-Event Fusion TAP'' (FE-TAP)~\cite{Liu24arxiv}, which implements correlation-based point tracking.
Event-only methods comparable to our~\mname~ comprise EM-ICP~\cite{Zhu17icra}, HASTE~\cite{Alzugaray20bmvc}, and DDFT E2VID~\cite{Messikommer23cvpr}, an adaptation of the combined method using E2VID- instead of grayscale-frames.
\Cref{tab:eds_ec_results} summarizes tracking results on the two datasets.
Our method outperforms all other event-based methods by a large margin (\gred{20}\% on EDS).
Remarkably, it also performs \gred{4.1}\% better than the best method using frames and events combined.

\def\figWidth{0.33\linewidth}
\begin{figure}[t]
	\centering
    {\footnotesize
    \setlength{\tabcolsep}{1pt}
	\begin{tabular}{
	>{\centering\arraybackslash}m{\figWidth} 
	>{\centering\arraybackslash}m{\figWidth} 
	>{\centering\arraybackslash}m{\figWidth} 
    }                                     %
		\gframe{\includegraphics[clip,trim={0cm 2cm 2cm 0cm},width=\linewidth]{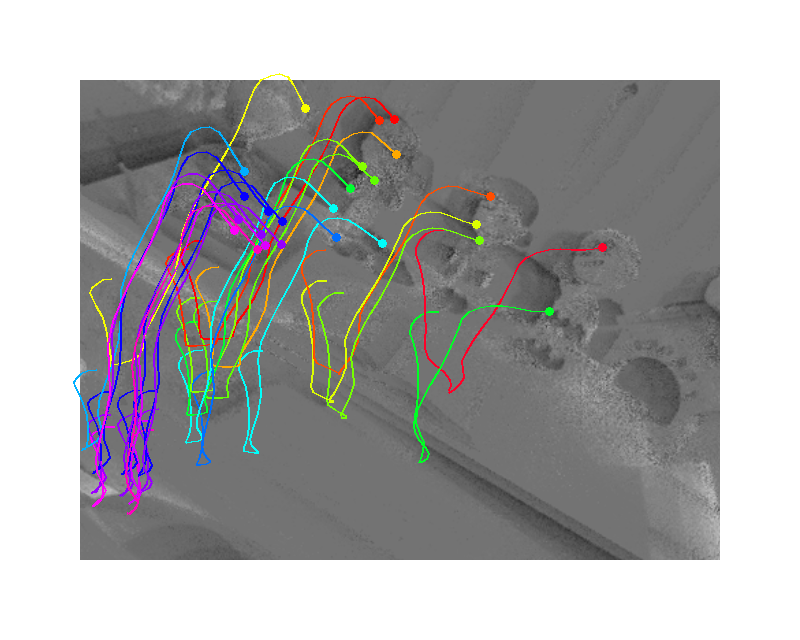}}
        &\gframe{\includegraphics[clip,trim={0cm 2cm 2cm 0cm},width=\linewidth]{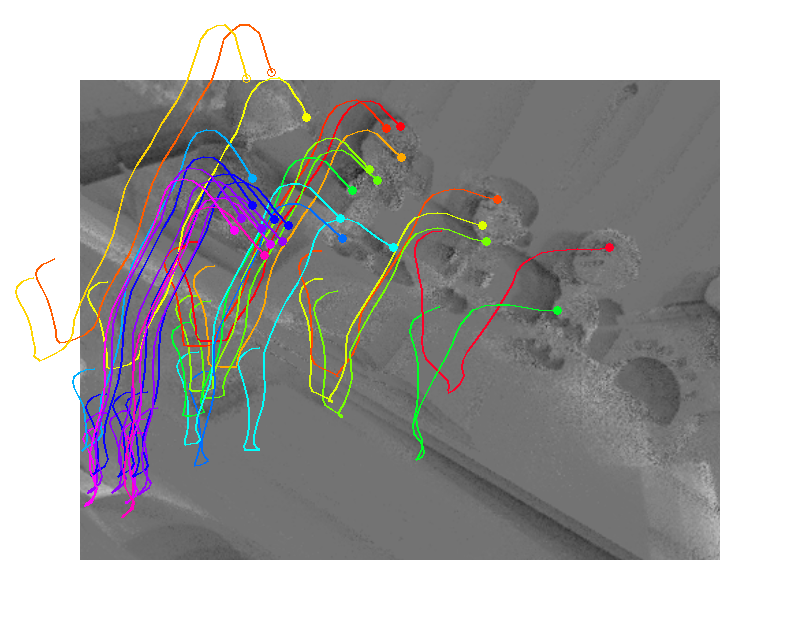}}
		&\gframe{\includegraphics[clip,trim={0cm 2cm 2cm 0cm},width=\linewidth]{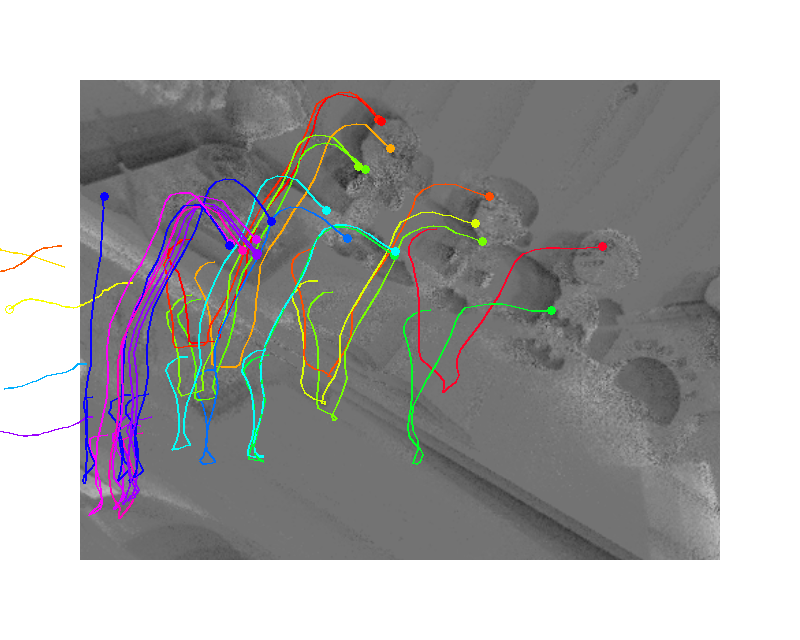}} \\
		GT
		& \textbf{Ours}
		& DDFT~\cite{Messikommer23cvpr}
	\end{tabular}
	}
    \vspace{-2ex}
	\caption{\emph{Task 2 - Feature tracking on EDS}.
Notably, our tracker captures points even after they leave the FOV and reenter.
    }
    \vspace{-1ex}
	\label{fig:exp:eds_ec}
\end{figure}
\begin{table}[t!]
\centering
\adjustbox{max width=\linewidth}{
\setlength{\tabcolsep}{4pt}
\begin{tabular}{lc*{4}{S[table-format=1.3,detect-weight,detect-mode]}}
\toprule
& & \multicolumn{2}{c}{EDS} & \multicolumn{2}{c}{EC} \\
\cmidrule(lr){3-4} \cmidrule(lr){5-6}
{Method} & {Input} & {\makecell{Feature\\Age $\uparrow$}} & {\makecell{Expected\\FA $\uparrow$}} & {\makecell{Feature\\Age $\uparrow$}} & {\makecell{Expected\\FA $\uparrow$}} \\
\midrule  %

{ICP~\cite{Kueng16iros}}  & E & 0.060 & 0.040 & 0.256 & 0.245 \\
{EKLT~\cite{Gehrig19ijcv}}            & E+F & 0.325 & 0.205 & 0.811 & 0.775 \\
{DDFT~\cite{Messikommer23cvpr}}       & E+F & 0.576 & 0.472  & 0.825 & 0.818 \\
{FE-TAP~\cite{Liu24arxiv}}            & E+F & 0.676 & 0.589  & 0.844 & 0.838  \\

\midrule  %
{EM-ICP~\cite{Zhu17icra}} & E & 0.161 & 0.120 & 0.337 & 0.334 \\
{HASTE~\cite{Alzugaray20bmvc}}    & E & 0.096 & 0.063 & 0.442 & 0.427 \\

DDFT E2VID~\cite{Messikommer23cvpr} & E & 0.589 & 0.495 & 0.794 & 0.786 \\

\textbf{\mname~w\textbackslash o FA-loss~(Ours)}    & E & \unum{1.3}{0.698} & \bnum{0.599} & \unum{1.3}{0.885} & \unum{1.3}{0.879} \\
\textbf{\mname~(Ours)}                              & E & \bnum{0.704} & \unum{1.3}{0.598} & \bnum{0.888} & \bnum{0.883} \\

\bottomrule
\end{tabular}
}
\vspace{-1ex}
\caption{\emph{Task 2 - feature tracking on EDS \& EC}. 
Input: E+F (Events \& Frames), E (Events only).
More results in Suppl.mat.
}
\vspace{-2ex}
\label{tab:eds_ec_results}
\end{table}
\begin{table*}[t!]
\centering
\adjustbox{max width=\textwidth}{
\setlength{\tabcolsep}{5pt}
\begin{tabular}{l*{7}{c}*{4}{S[table-format=1.3,detect-weight,detect-mode]}}
\toprule
& & & & & & & & \multicolumn{2}{c}{EDS} & \multicolumn{2}{c}{EC} \\
\cmidrule(lr){9-10} \cmidrule(lr){11-12}
{Name} & 
{Resolution} & 
{\makecell{Contrast\\thresholds}} & 
{Augment} & 
{Dataset} & 
{\makecell{Base\\fps}} & 
{\makecell{Varying\\dynamics}} & 
{\makecell{DNN\\input}} & 
{\makecell{Feature\\Age $\uparrow$}} & 
{\makecell{Expected\\FA $\uparrow$}} & 
{\makecell{Feature\\Age $\uparrow$}} & 
{\makecell{Expected\\FA $\uparrow$}} \\
\midrule %
\textbf{Baseline}           &  %
(256,256)                   &  %
{0.2}                       &  %
{--}                        &  %
{MOVI-F}                    &  %
{12}                        &  %
{\xmark}                    &  %
{event stack}               &  %
0,598                       &  %
0,500                       &  %
0.780                       &  %
0.775                       \\ %
\midrule %
{High resolution}    &  %
(512,512)                   &  %
{0.2}                       &  %
{--}                        &  %
{MOVI-F}                    &  %
{12}                        &  %
{\xmark}                    &  %
{event stack}               &  %
\bnum{0.659}                &  %
\bnum{0,561}                &  %
\unum{1.3}{0.808}           &  %
\unum{1.3}{0.802}           \\ %
{Random Thresholds as in ~\cite{Klenk24threedv}}         &  %
(256,256)                   &  %
$\sim\mathcal{U}(0.16, 0.34)$   &  %
{--}                        &  %
{MOVI-F}                    &  %
{12}                        &  %
{\xmark}                    &  %
{event stack}               &  %
0.627                       &  %
0.531                       &  %
\bnum{0.836}                &  %
\bnum{0.830}                \\ %
{Random Thresholds as in ~\cite{Stoffregen20eccv}}         &  %
(256,256)                   &  %
$\sim\mathcal{U}(0.20, 1.50)$     &  %
{--}                        &  %
{MOVI-F}                    &  %
{12}                        &  %
{\xmark}                    &  %
{event stack}               &  %
0.609                       &  %
0.519                       &  %
0.801                       &  %
0.795                       \\ %
{Frame Rate Influence}        &  %
(256,256)                     &  %
{0.2}                         &  %
{--}                          &  %
{MOVI-F + \dname$_{static}$}  &  %
{12 - 48}                     &  %
{\xmark}                      &  %
{event stack}                 &  %
0.618                              &  %
0.514                              &  %
0.781                              &  %
0.777                              \\ %
{Varying Dynamics Influence}        &  %
(256,256)                     &  %
{0.2}                      &  %
{--}                         &  %
{MOVI-F + \dname$_{dynamic}$}    &  %
{12 - 48}                    &  %
{\cmark}                    &  %
{event stack}                    &  %
0.617                           &  %
0.528                           &  %
0.781                           &  %
0.776                           \\ %
{Noise Augmentation}        &  %
(256,256)                   &  %
{0.2}                      &  %
{Gauss. noise}              &  %
{MOVI-F}                    &  %
{12}                        &  %
{\xmark}                    &  %
{event stack}                    &  %
0.631                       &  %
0.530                       &  %
0.822                       &  %
0.816                       \\ %
{Representation Influence}  &  %
(256,256)                   &  %
{0.2}                      &  %
{--}                        &  %
{MOVI-F}                    &  %
{12}                        &  %
{\xmark}                    &  %
{voxel grid}                &  %
0.592                       &  %
0.505                       &  %
0.805                       &  %
0.799                       \\ %
{MultiFlow}                 &  %
(512,384)                   &  %
{--}                        &  %
{--}                        &  %
{MultiFlow~\cite{Gehrig24pami}}                    &  %
{1000}                      &  %
{N/A}                       &  %
{event stack}               &  %
0.221                       &  %
0.178                       &  %
0.323                       &  %
0.316                       \\ %
\textbf{\dname~(Ours)}                &  %
(256,256)                      &  %
$\sim\mathcal{U}(0.16, 0.34)$  &  %
{--}                           &  %
{\dname~(Ours)}                &  %
{48}                           &  %
{\cmark}                       &  %
{event stack}                  &  %
\unum{1.3}{0.646}              &  %
\unum{1.3}{0.550}              &  %
0.777                          &  %
0.772                          \\ %
\bottomrule
\end{tabular}
}
\vspace{-1ex}
\caption{\emph{Sensitivity study}. Analysis of different parameter configurations and their impact on feature tracking performance. Higher values are better (↑).
Note that most models were trained on a smaller resolution than the final model ($256 \times 256$ px). 
}
\label{tab:sensitivity-study}
\vspace{-2ex}
\end{table*}

\Cref{fig:exp:eds_ec} shows that our method tracks points precisely and even further recovers tracks well when they leave the frame and reenter, which is often not captured by the ground truth and not reflected in the metrics.

\subsection{Sensitivity and Ablations}
\label{sec:exp:sensitivity}

\subsubsec{Contrastive loss.}
Our final model was trained for $10^5$ steps without the contrastive loss and further refined for $1.2 \cdot 10^5$ steps including it.
We continue to train a comparison model without contrastive loss from the same checkpoint and provide results on all datasets (\cref{tab:eds_ec_results,tab:results_event_kubric,tab:e2d2_fidget_results}) marked as \mname~w\textbackslash o FA-loss~(Ours).
The loss gives a slight boost across all real event datasets (except the synthetic {\dname}) and helps the network to learn motion-robust features.

\subsubsec{Dataset and Training Sensitivity.}
We provide the results of sensitivity studies conducted during dataset construction.
In preparation of our self-rendered dataset, we created baseline datasets using MOVi-F, a freely available pre-rendered Kubric dataset.
We applied the same event and point track generation procedures as in~\cref{sec:data} with different versions for different contrast sensitivities.
We run Vid2e on 512x512px resolution but downsample at training time for most experiments.
The datasets served as a development benchmark to validate design choices.
All models were trained for $1.7 \cdot 10^5$ training steps on four Nvidia A6000 GPUs with an effective batch size of 8 (except high-res. on A100).
\cref{tab:sensitivity-study} shows the conducted experiments.
We used metrics on EDS for design decisions as it is more consistent than EC, where metrics often alternate between epochs.

The results show the biggest improvement for higher resolution and for choosing random thresholds for event generation of $\sim\mathcal{U}(0.16, 0.34)$ as reported in~\cite{Klenk24threedv}.
The column \textit{base fps} indicates the rendered framerate (before FILM upsampling).
The influence of frame rate and varying dynamics were measured before rendering a whole dataset and therefore tested with only 3500 samples, respectively, and paired with samples of MOVi-F to match the number of baseline samples for comparability.
We found both measures (increasing the base frame rate from 12 to 48 and using panning motion) effective, increasing the performance by $\approx$2\% each.
The results confirm the effectiveness of the Gaussian noise augmentation and show a slight performance advantage for event stacks over voxel grids.
Lastly, we trained our method on MultiFlow.
It does not provide meta information to derive visibility flags, which we set to 1 for all tracks.
In our tests, we only were able to achieve inferior results compared to training on {\dname}.

\subsection{Limitations}
\label{sec:limitations}

Since high-resolution event cameras only provide monochrome information, they cannot yet leverage color information to establish appearance correspondences between points.
Furthermore, we observe that our method relies on query times during scene motion.Track features $Q_t^i$ initialized in the absence of motion, and therefore events, do not capture the scene appearance well.
This is an inherent problem of event data and could be addressed by reinitializing track features as soon as motion is detected.

\section{Conclusion}
\label{sec:conclusion}

We introduce the first event-only method for tracking any point in a data stream.
The method shows strong performance on five datasets, across different camera types and resolutions, and outperforms all compared methods on a common feature tracking benchmark by a large margin.
Its capability is driven by the rigorous design of a new synthetic dataset and a contrastive loss providing robustness of correlation features.
Results also show scenarios where our event-only method has advantages over frame-based ones.

\section*{Acknowledgements}
We thank Dr. Fermüller and NeuroPAC for fostering collaborations within the event-based community (NSF OISE 2020624).
Funded by the DFG (German Research Foundation) – EXC 2002/1 “Science of Intelligence” – project no. 390523135.
We furthermore gratefully acknowledge the support of the following grants: NSF FRR 2220868, NSF IIS-RI 2212433, NSF TRIPODS 1934960, ONR N00014-22-1-2677, NSF NCS-FO 2124355, SNF 225354.

\ifarxiv
\section*{Supplementary Material}
\else
\clearpage
\setcounter{page}{1}
\maketitlesupplementary
\fi

\section{Method Details}

\begin{figure}[t]
   \centering
   {\includegraphics[trim={9.2cm 6.5cm 15.6cm 10.3cm},clip,width=.9\linewidth]{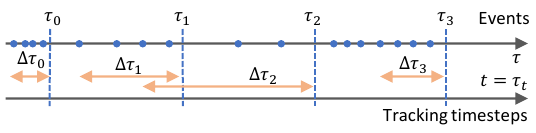}}
   \label{fig:suppl:event_repr}
   \caption{Asynchronous events are converted into temporally equidistant frame representations at $\tau_t$, each created from the last $N_e$ events.}
\end{figure}

\subsubsec{Clarifications Event Representation}
Events are quasi-continuous.
\Cref{eq:tracker} defines the task of tracking any point from events as determining the time-discrete point observations from the continuous input events.
In the first step events are converted to event representations, where each representation has a constant number of events $N_e$.
\Cref{fig:suppl:event_repr} shows exemplary the connection between events and discrete tracking timesteps $\tau$, resulting in a constant tracking frequency, despite a varying event rate.
Please note that the tracking frequency is adjustable at test time.
In practice, we mostly set $\tau_t$ to the ground truth timesteps of an evaluation set.

\subsubsec{Description of Event Stacks}
As frame representation, we use a variation of Mixed-Density event stacks \cite{Nam22cvpr} and build $T$ input representations $I_t$.
Let $E_t=\{e_i|t_i \leq \tau_t\}$ be the $N_e$ events directly preceding timestep $\tau_t$. We construct a multi-channel representation by hierarchically binning these events into $C = 10$ channels, denoted as $\{h_c\}_{c=1}^C$, where each channel $h_c$ is a spatial histogram of dimensions $H \times W$.
The $c$-th channel aggregates $n_c = \lfloor N_e/2^{c-1} \rfloor$ events using bilinear interpolation, such that:
\begin{itemize}
    \item $h_1$ incorporates all $N_s$ events
    \item $h_c$ processes $N_s/2^{c-1}$ events for $c > 1$
\end{itemize}
where each channel contains the events closest to $t_i$.

\subsubsec{Hyperparameters}
For a better overview \cref{tab:model_hparams} provides an overview of all hyperparameters of our method introduced in \cref{sec:method}.

\subsubsec{Event Generation Model}
The linear event generation model has been discussed previously (e.g.~\cite{Gallego20pami}.
To make the paper self-contained, here is a brief introduction.
It approximates how events are triggered in event cameras.
Starting from the condition that events occur when brightness change reaches a threshold ($\Delta \Lum(\bx_k,t_k) = \pol_k \,C$), this model uses Taylor's expansion for small time intervals to relate events to the temporal derivative of brightness ($\Delta{\Lum}{t}(\bx_k,t_k) \approx \frac{\pol_k\, C}{\Delta t_k}$).
Under constant illumination, this can be further linearized to $\Delta \Lum \approx - \nabla \Lum \cdot v \Delta t$, showing that events are fundamentally triggered by brightness gradients (edges) moving across the image plane.
The rate of event generation depends on the relationship between edge orientation and motion direction, with perpendicular motion producing the highest event rate.

\subsubsec{Events under Time Inversion.}
According to the linearized event generation model (LEGM)~\cite{Gallego20pami} an event $e_k$ is generated when the dot product between per-pixel optical flow $v$ and the image gradient $\nabla \Lum$ exceeds the threshold $C$:
\begin{equation}
    \label{eq:eventgen}
    e_k \in E_t \iff -p_k \nabla \Lum(\bx_k,\tau_k) \cdot v(\bx_k,\tau_k)\delta\tau_k  \approx C
\end{equation}
where $\delta \tau_k$ is the time since the last event at the same pixel.

Next, consider how the events $E_t$ change when the motion changes,
for example, induced by a time inversion $\tilde{\tau} \doteq 2\bar{\tau}_t - \tau$, with $\bar{\tau}_t=\frac{\tau_t + \tau_t - \Delta \tau_t}{2}$ is the interval midpoint.
Due to the chain rule, the optical flow becomes $\tilde{v}(\bx,\tau)=-v(\bx,2\bar{\tau}_t-\tau)$, and the gradient becomes $\nabla \tilde{\Lum}(\bx,\tau)=\nabla \Lum(\bx,2\bar{\tau}_t - \tau)$.
Under this change of variables, we describe what the new events $\tilde{E}_t$ look like.
Specifically, if $e_k\in E_t$, then $\tilde{e}_k=(\bx_k,2\bar{\tau}_t-\tau_k,-p_k)\in \tilde{E}_t$ since 
\begin{align}
    &-\tilde{p}_k \nabla \tilde{\Lum}(\tilde{\bx}_k,\tilde{\tau}_k) \cdot \tilde{v}(\tilde{\bx}_k,\tilde{\tau}_k)\delta\tilde{\tau}_k\\
    \nonumber&=-p_k \nabla \Lum(\bx_k,\tau_k) \cdot v(\bx_k,\tau_k)\delta\tau_k \stackrel{\eqref{eq:eventgen}}{\approx} C. \end{align}
The equality is satisfied assuming the time since the last event is similar under time inversion ($\delta \tilde{\tau}_k\approx \delta \tau_k$).
Simple inspection shows that the events $E_t$ and $\tilde{E}_t$ are different, and, as a result, corresponding descriptors $D_t^s$ and $\tilde{D}_t^{w-s+1}$ are different (note $w-s+1$ is the inverted index).

\section{Data and Evaluation Details}
\begin{table}[t!]
\centering
\adjustbox{max width=0.7\linewidth}{
\setlength{\tabcolsep}{3pt}
\begin{tabular}{lcr}
\toprule
Parameter & Variable & Value \\
\midrule
window length            & $w$   & 8 \\
feature size             & $d$   & 128 \\
bin number               & $B$   & 10 \\
stride                   & $T_s$ & 4 \\
refinement steps (train) & $M$   & 4 \\
refinement steps (eval)  & $M$   & 6 \\
feature scales           & $S$   & 4 \\
\bottomrule
\end{tabular}
}
\vspace{-1ex}
\caption{\emph{Hyperparameters}.
An overview of variables that were introduced in \cref{sec:method} and their specific values.
}
\label{tab:model_hparams}
\end{table}

\subsection{Ground truth generation for the E2D2 Fidget Spinner Sequence}

\begin{figure}[t]
   \centering
   \begin{subfigure}[b]{0.4\linewidth}
      \centering
      {\includegraphics[trim={9cm 5.5cm 6cm 4cm},clip,width=\linewidth]{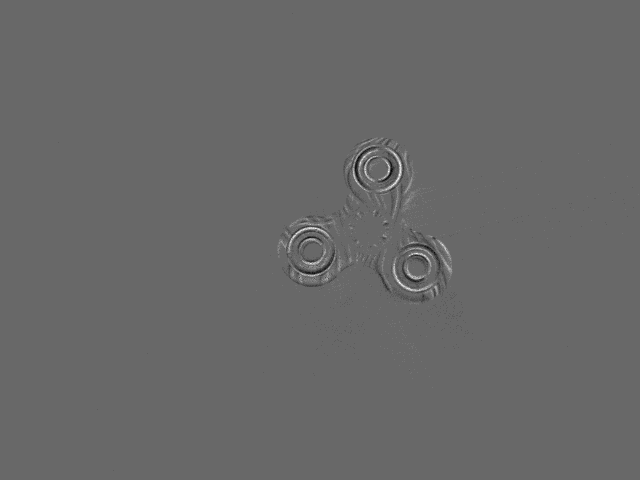}}
      \caption{}
   \end{subfigure}
   
   \begin{subfigure}[b]{.8\linewidth}
      \centering
      {\includegraphics[trim={0cm 0cm 0cm 0cm},clip,width=\linewidth]{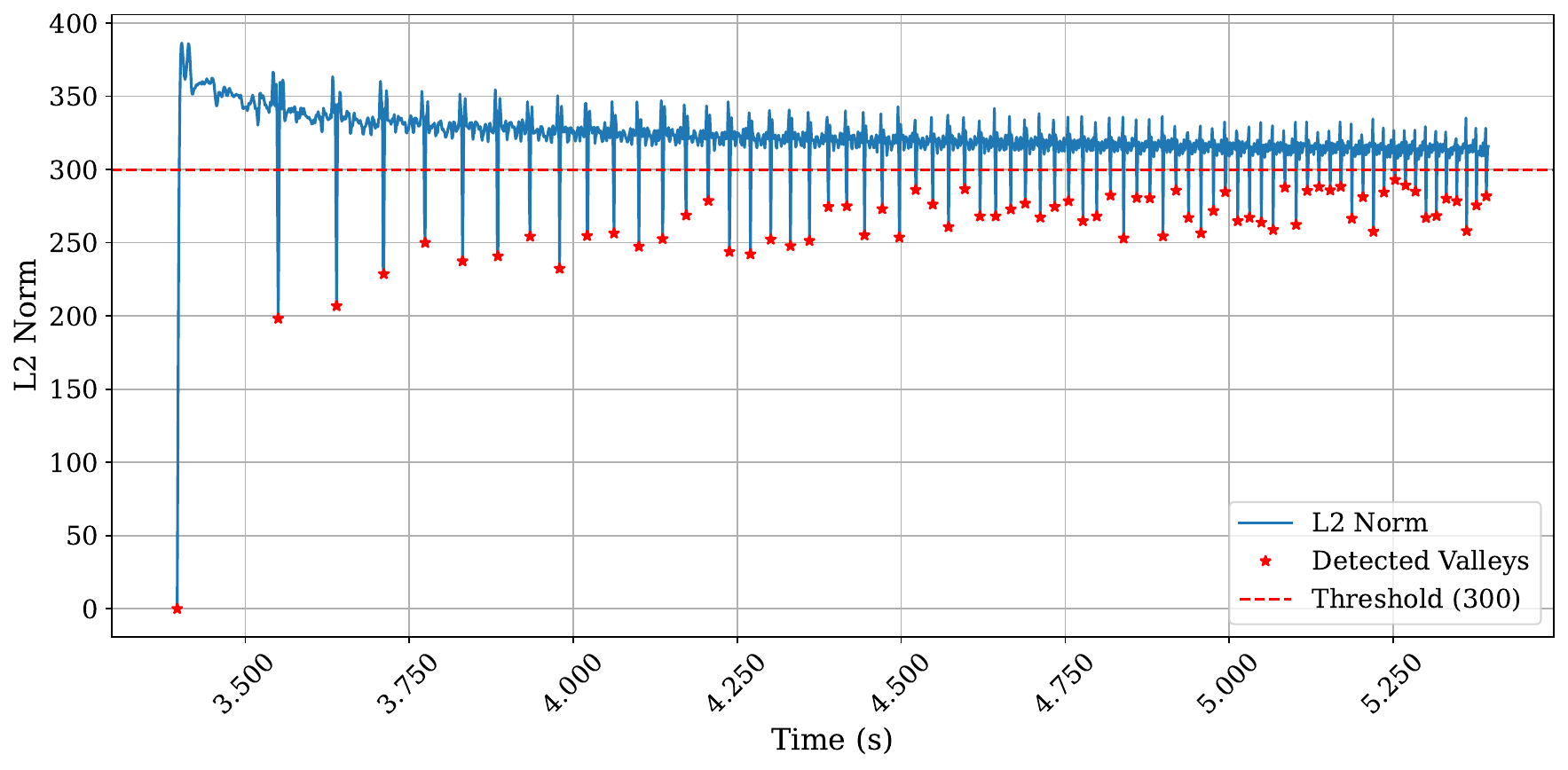}}
      \caption{}
   \end{subfigure}
\caption{\emph{Ground truth for E2D2 fidget spinner sequence}. (a) Example of a 2D event histogram that is built at 1000Hz. (b) time series of L2 norms wrt. to the first frame.
Red star points are local minima, where the spinner completed another third revolution.}
\label{fig:e2d2_gt_l2norms}
\end{figure}

The ground truth tracks used for evaluation on the E2D2 fidget spinner sequence were calculated from simple geometric knowledge.
The midpoint of the spinner is constant. The wheel itself is fully facing the camera, describing perfect circular motions.
Therefore, we can calculate the positions of each point on the fidget spinner with an estimate of the angular velocity of the wheel.
The angular velocity is estimated as follows:
First, we create event histograms with a fixed number of 20,000 events at 1000 Hz (simply counting positive and negative events within the event batch), as seen in \Cref{fig:e2d2_gt_l2norms} (a).
Then we calculate the 1D time series of the L2-norm between each frame and the initial frame, visualized in \cref{fig:e2d2_gt_l2norms} (b).
The local minima are the times when the wheel completed a third revolution (due to the three-lobed shape of the fidget spinner).
We assume the angular velocity to be constant between two third-revolution-timestamps.
As shown in \cref{fig:e2d2_gt_l2norms} (b), the spinner gets progressively faster, increasing tracking difficulty.

\def\figmethodwidth{.5\linewidth}

\subsection{Examples of the EventKubric Dataset}

\begin{figure}[t]
   \centering
   {\includegraphics[trim={7.5cm 2cm 7.5cm 3.5cm},clip,width=\linewidth]{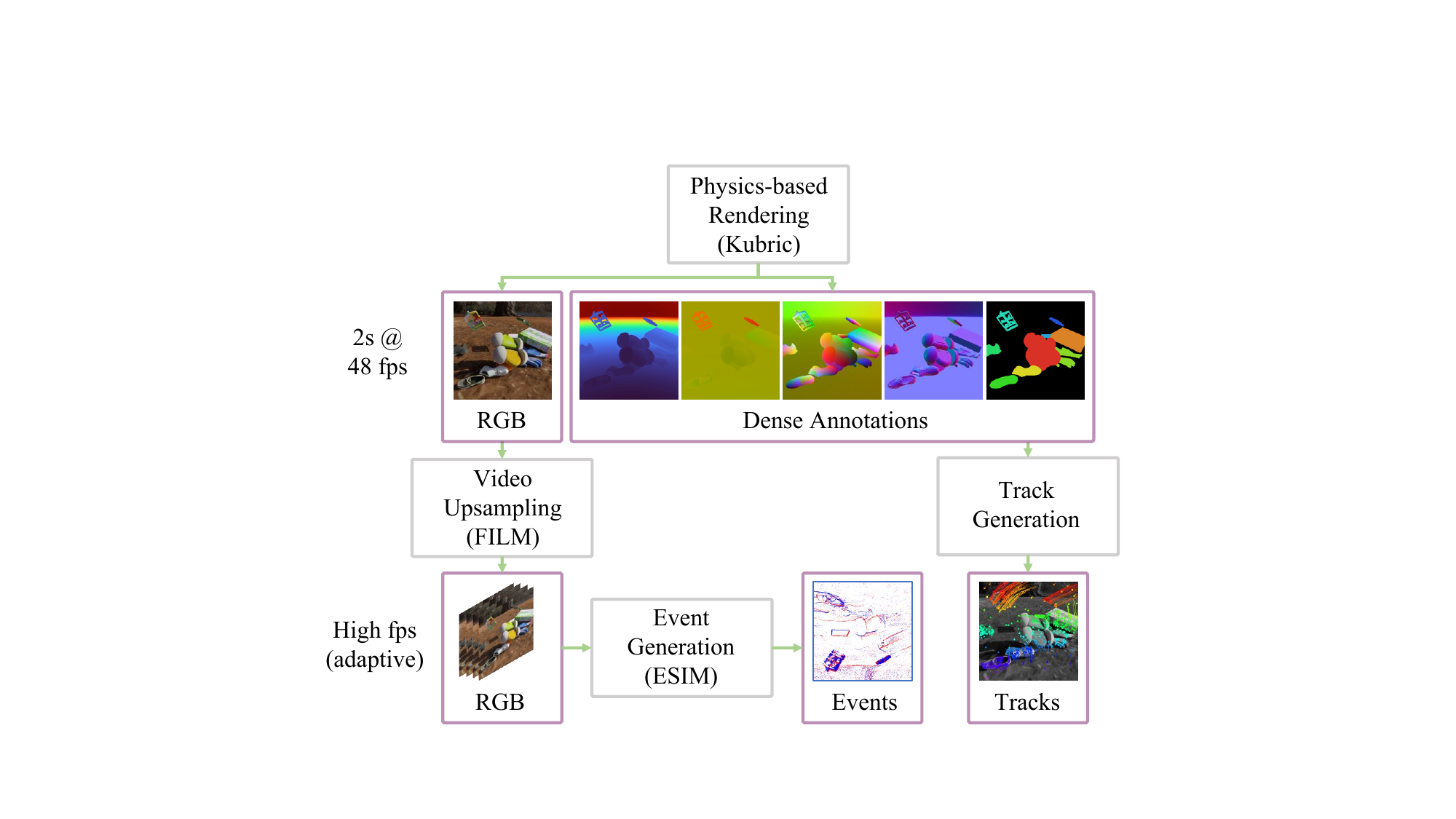}}
\caption{\emph{Data Generation Pipeline}.
The PBR tool Kubric renders 2s RGB videos, which are adaptively upsampled to generate events from it. The dense ground truth provided by Kubric is used for point track generation.}
\label{fig:data_pipeline}
\end{figure}

\Cref{fig:data_pipeline} visualizes the data generation explained in \cref{sec:data}.
\Cref{fig:data:examples} shows a few examples of the {\dname} dataset.
The full scene knowledge is available as annotations, which can be useful for tasks beyond point tracking.

\def\figmethodwidth{.48\linewidth}

\section{Further Experiments and Detailed Results}

\subsection{Task 2: Feature Tracking - Extended Results}

\Cref{tab:detailed-metrics} provides full results for the EDS \& EC dataset.
\Cref{fig:exp:eds_ec_big} shows additional comparisons.

\subsection{Results EVIMO2}

\Cref{fig:evimo_pred} shows prediction results for EVIMO2

\subsection{Feature Independence Experiment.}

\begin{figure}[t]
   \centering
   {\includegraphics[trim={4cm 6cm 8cm 7.5cm},clip,width=.9\linewidth]{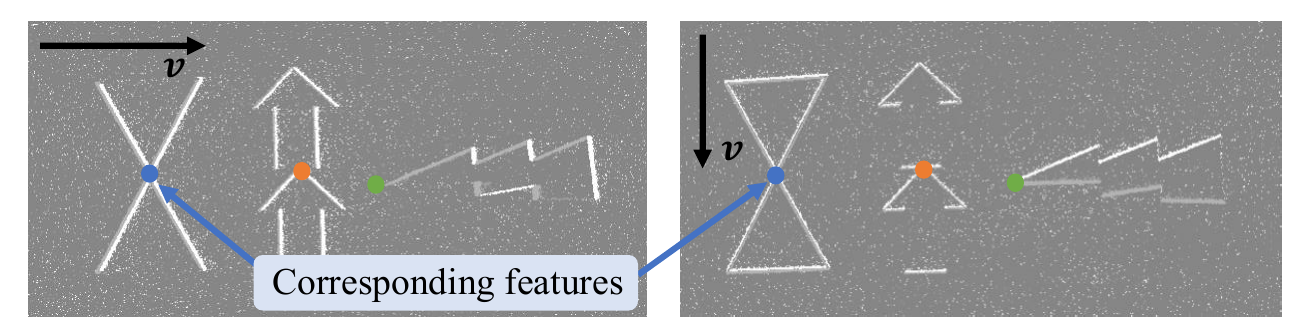}}
\caption{\emph{Setup of the motion robustness experiment}.
The same pattern is recorded two times in perpendicular directions at the same key points of the pattern.
The same points under different motion directions should ideally have similar descriptors.
}
\label{fig:feature_inv_exp}
\vspace{-2ex}
\end{figure}

We examine the effect of our contrastive loss on the learned features with an experiment shown in \cref{fig:feature_inv_exp}.
We track the same 3 points on a 2D pattern with two orthogonal camera motions and analyze the corresponding descriptors $d^{i}_{t,\text{dir}}$ at the end of the window with point index $i$ and $\text{dir}\in \{\text{horizontal},\text{vertical}\}$.
We then measure the cosine similarity between descriptors at the trajectory start, and descriptors along the same trajectory with $\mathcal{C}_\text{intra} = \sum_{t,\text{dir},i} \text{cos}_\text{sim}(d^{i}_{0,\text{dir}}, d^{i}_{t,\text{dir}})$, called \textit{intra-cluster}, and along trajectories with \emph{different motions directions} e.g. $\mathcal{C}_\text{inter} = \sum_{t,i} \text{cos}_\text{sim}(d^{i}_{0,\text{horizontal}}, d^{i}_{t,\text{vertical}})$, called \textit{inter-cluster}.
\Cref{tab:feature_independence} shows results for three methods: our model, an ablation model trained without our loss, and a frame-based baseline.
While the model in the motion-independent frame domain has very similar inter- and intra-cluster similarities, the ablation model shows a similarity gap of 0.38 between $\mathcal{C}_\text{intra}$ and $\mathcal{C}_\text{inter}$.
In comparison, this gap is closed, when training with our contrastive loss.

\def\figmethodwidth{.5\linewidth}
\begin{table}[t!]
\centering
\adjustbox{max width=0.7\linewidth}{
\setlength{\tabcolsep}{3pt}
\begin{tabular}{l*{3}{S[table-format=1.3,detect-weight,detect-mode]}}
\toprule
{Method} & {\makecell{$\mathcal{C}_\text{intra} \uparrow$}} & {\makecell{$\mathcal{C}_\text{inter} \uparrow$}} & $\Delta$ \\
\midrule
{Frames}                   & 0.836 & 0.804 & \textbf{0.032} \\
{Events without FA-loss}   & 0.776 & 0.399 & 0.377 \\
{Events with FA-loss}      & \textbf{0.954} & \textbf{0.887} & 0.067 \\
\bottomrule
\end{tabular}
}
\caption{\emph{Measuring feature independence}.
The intra- and inter-cluster cosine similarity of tracking the same points in different sequences.
}
\label{tab:feature_independence}
\end{table}

\def\figWidth{0.15\linewidth}
\begin{figure*}[t]
    \centering
    {\scriptsize
    \setlength{\tabcolsep}{1pt}
    \begin{tabular}{
    >{\centering\arraybackslash}m{1cm} 
    >{\centering\arraybackslash}m{\figWidth}
    >{\centering\arraybackslash}m{\figWidth}
    >{\centering\arraybackslash}m{\figWidth}
    >{\centering\arraybackslash}m{\figWidth}
    }
        \rotatebox{90}{\textbf{(a)} Events}
        &\gframe{\includegraphics[clip,trim={0cm 0cm 0cm 0cm},width=\linewidth]{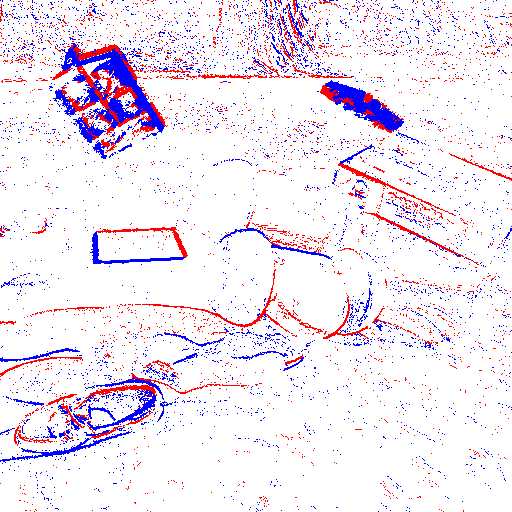}}
        &\gframe{\includegraphics[clip,trim={0cm 0cm 0cm 0cm},width=\linewidth]{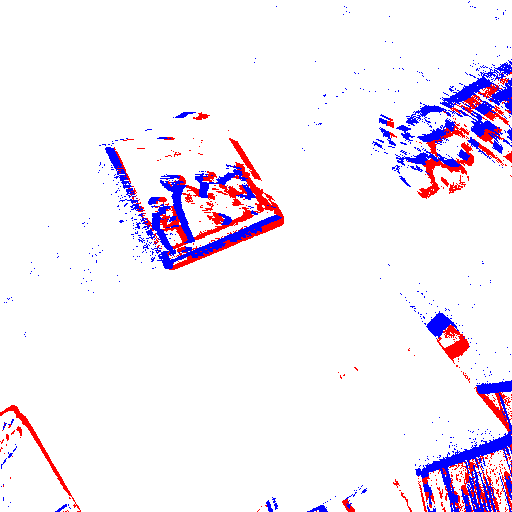}}
        &\gframe{\includegraphics[clip,trim={0cm 0cm 0cm 0cm},width=\linewidth]{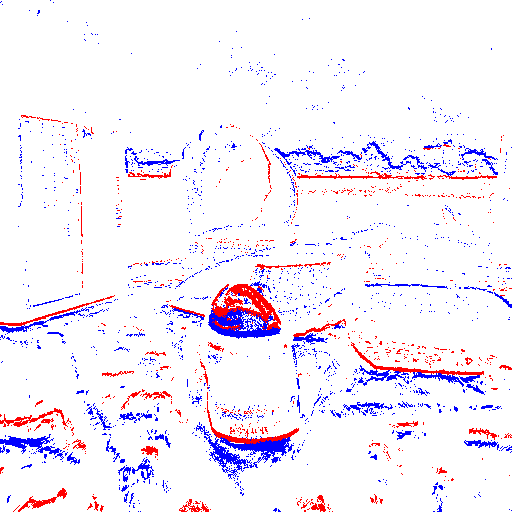}}
        &\gframe{\includegraphics[clip,trim={0cm 0cm 0cm 0cm},width=\linewidth]{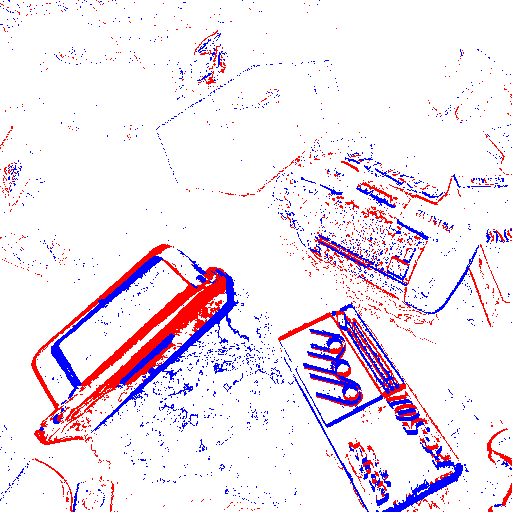}} \\
        \rotatebox{90}{\textbf{(b)} RGB}
        &\gframe{\includegraphics[clip,trim={0cm 0cm 0cm 0cm},width=\linewidth]{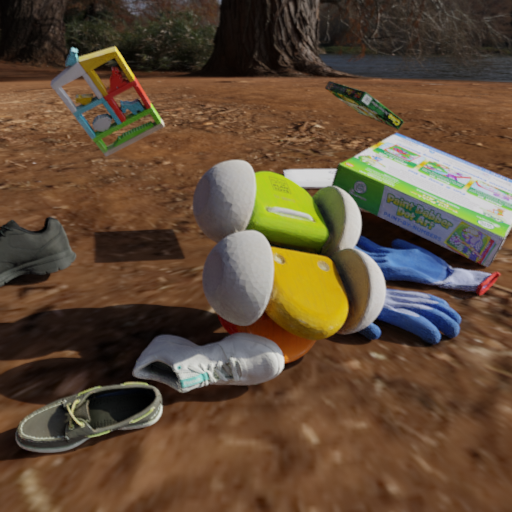}}
        &\gframe{\includegraphics[clip,trim={0cm 0cm 0cm 0cm},width=\linewidth]{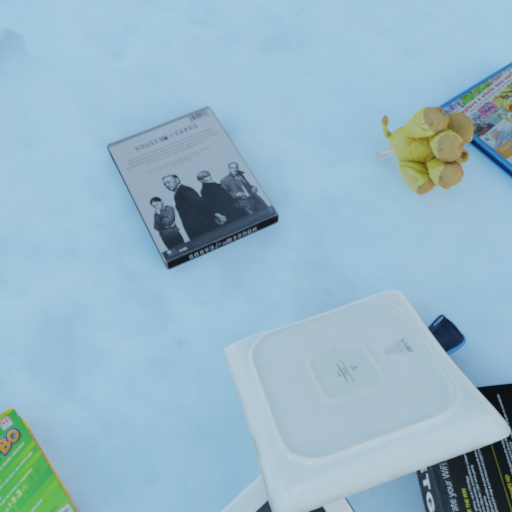}}
        &\gframe{\includegraphics[clip,trim={0cm 0cm 0cm 0cm},width=\linewidth]{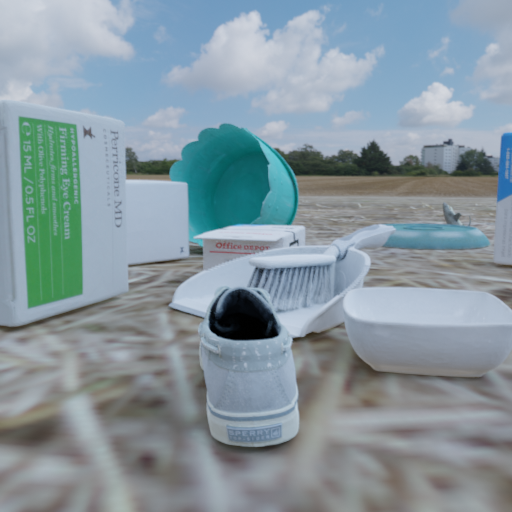}}
        &\gframe{\includegraphics[clip,trim={0cm 0cm 0cm 0cm},width=\linewidth]{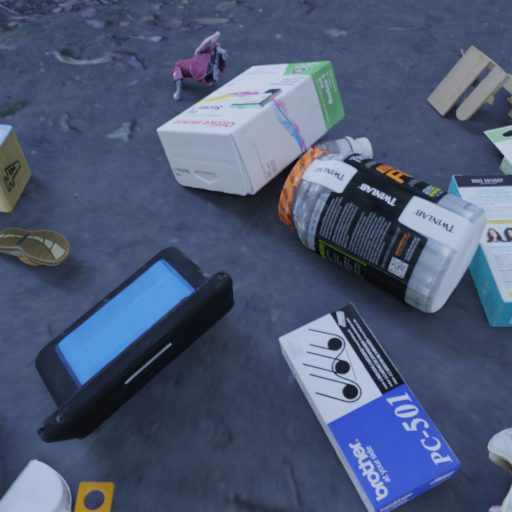}} \\
        \rotatebox{90}{\textbf{(c)} Point Tracks}
        &\gframe{\includegraphics[clip,trim={0cm 0cm 0cm 0cm},width=\linewidth]{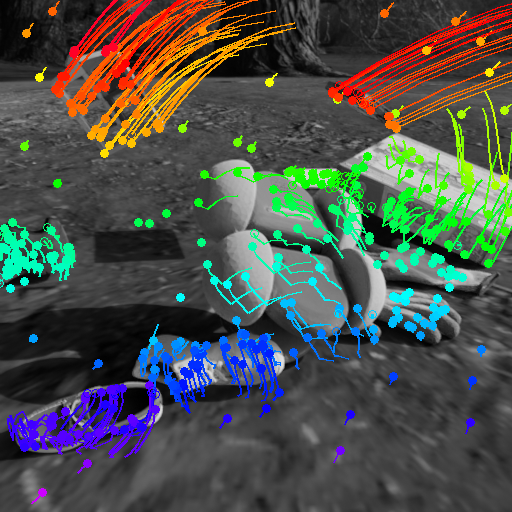}}
        &\gframe{\includegraphics[clip,trim={0cm 0cm 0cm 0cm},width=\linewidth]{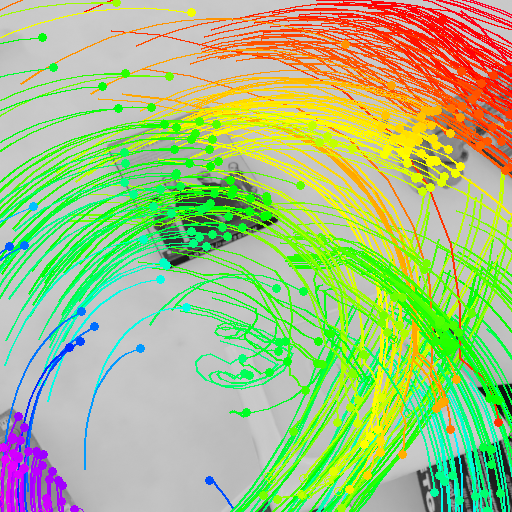}}
        &\gframe{\includegraphics[clip,trim={0cm 0cm 0cm 0cm},width=\linewidth]{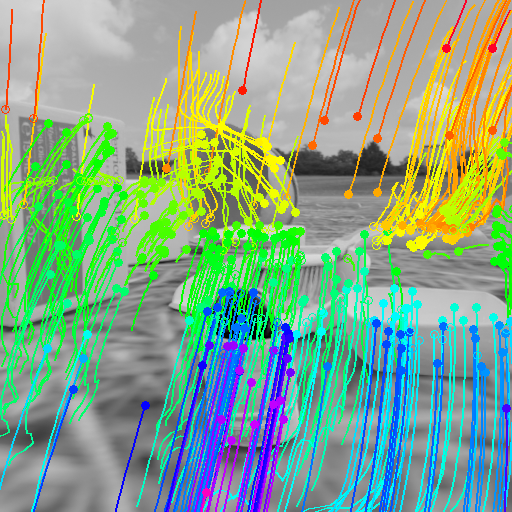}}
        &\gframe{\includegraphics[clip,trim={0cm 0cm 0cm 0cm},width=\linewidth]{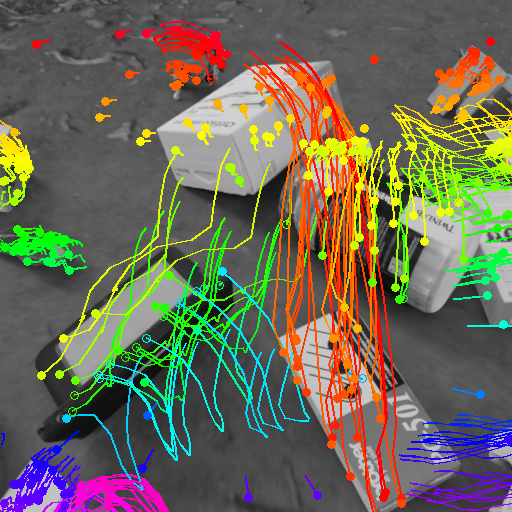}} \\
        \rotatebox{90}{\textbf{(d)} Forward Flow}
        &\gframe{\includegraphics[clip,trim={0cm 0cm 0cm 0cm},width=\linewidth]{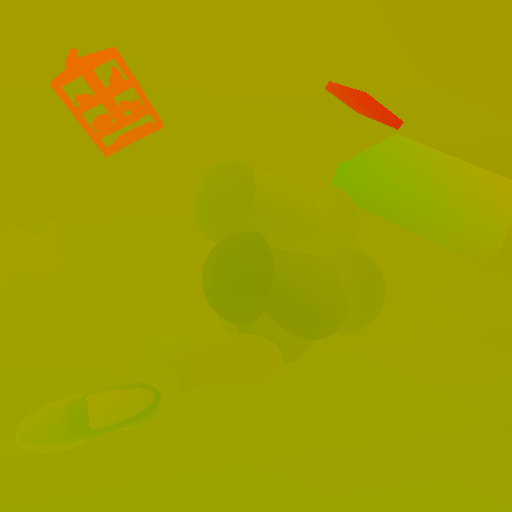}}
        &\gframe{\includegraphics[clip,trim={0cm 0cm 0cm 0cm},width=\linewidth]{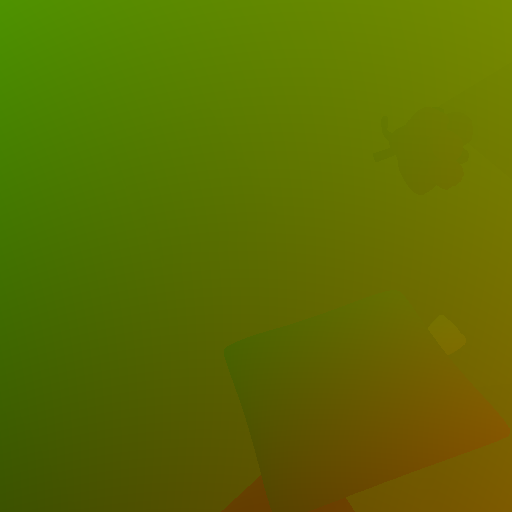}}
        &\gframe{\includegraphics[clip,trim={0cm 0cm 0cm 0cm},width=\linewidth]{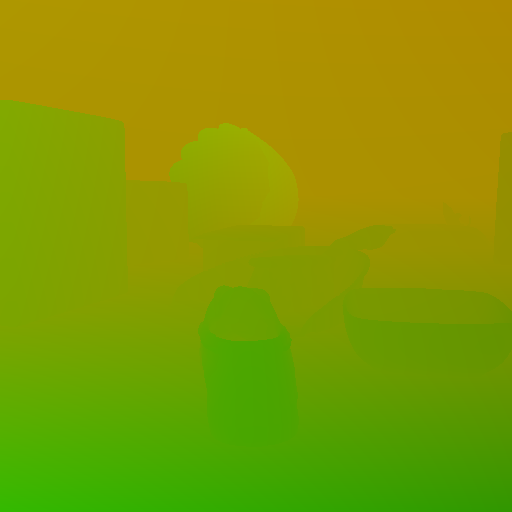}}
        &\gframe{\includegraphics[clip,trim={0cm 0cm 0cm 0cm},width=\linewidth]{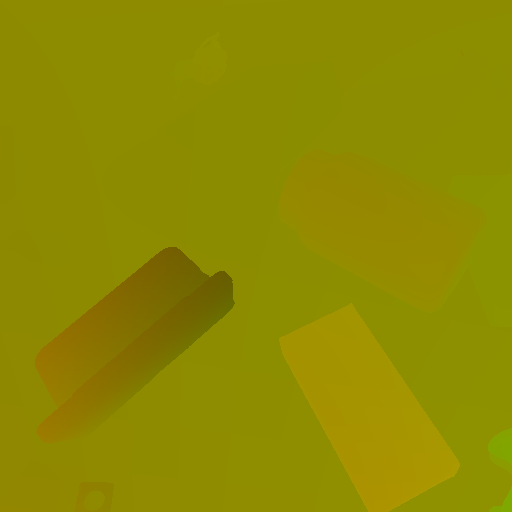}} \\
        \rotatebox{90}{\textbf{(e)} Depth}
        &\gframe{\includegraphics[clip,trim={0cm 0cm 0cm 0cm},width=\linewidth]{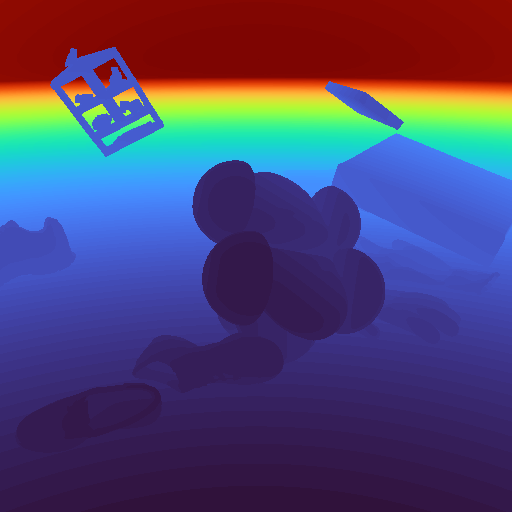}}
        &\gframe{\includegraphics[clip,trim={0cm 0cm 0cm 0cm},width=\linewidth]{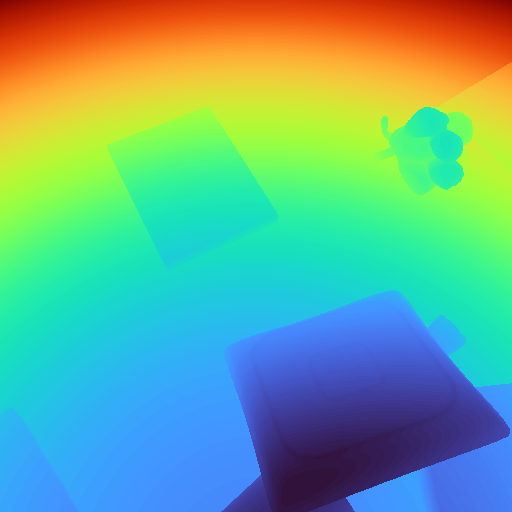}}
        &\gframe{\includegraphics[clip,trim={0cm 0cm 0cm 0cm},width=\linewidth]{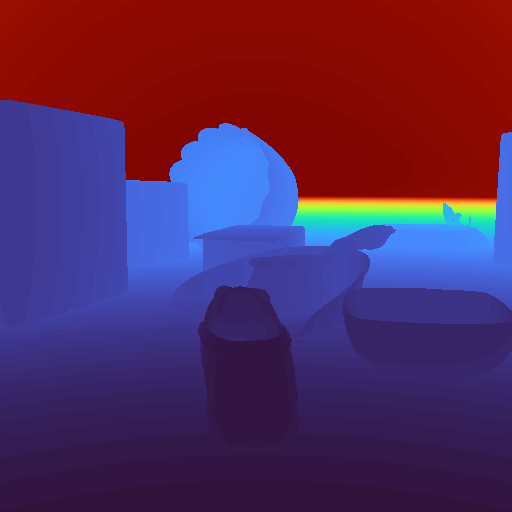}}
        &\gframe{\includegraphics[clip,trim={0cm 0cm 0cm 0cm},width=\linewidth]{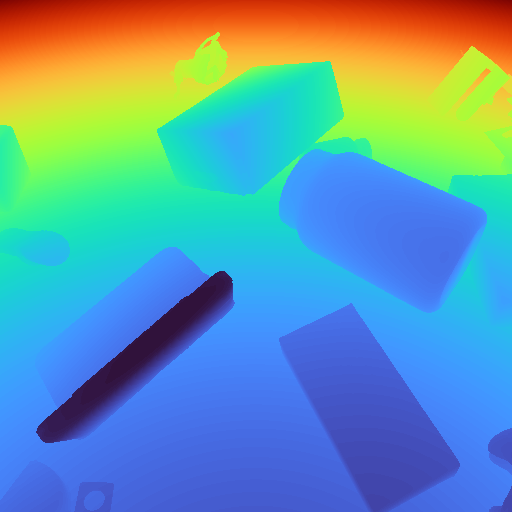}} \\
        \rotatebox{90}{\textbf{(f)} Surface Normal}
        &\includegraphics[clip,trim={0cm 0cm 0cm 0cm},width=\linewidth]{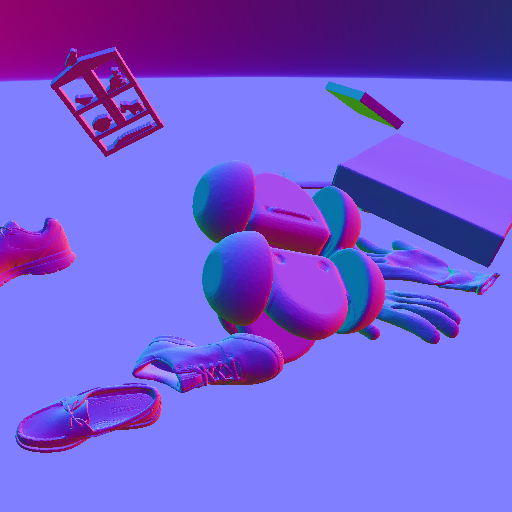}
        &\includegraphics[clip,trim={0cm 0cm 0cm 0cm},width=\linewidth]{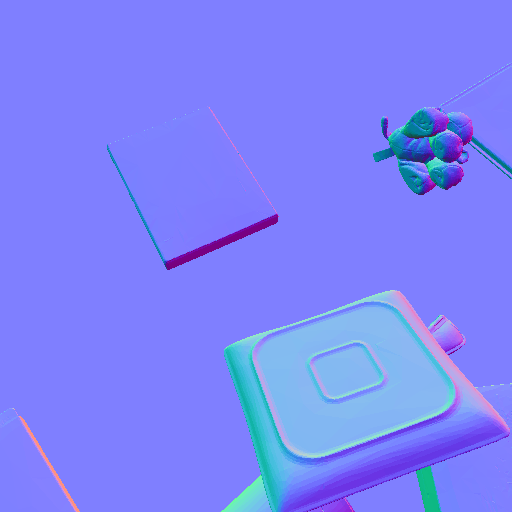}
        &\includegraphics[clip,trim={0cm 0cm 0cm 0cm},width=\linewidth]{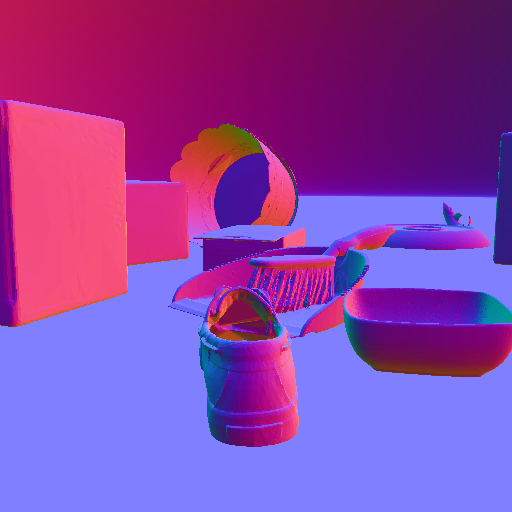}
        &\includegraphics[clip,trim={0cm 0cm 0cm 0cm},width=\linewidth]{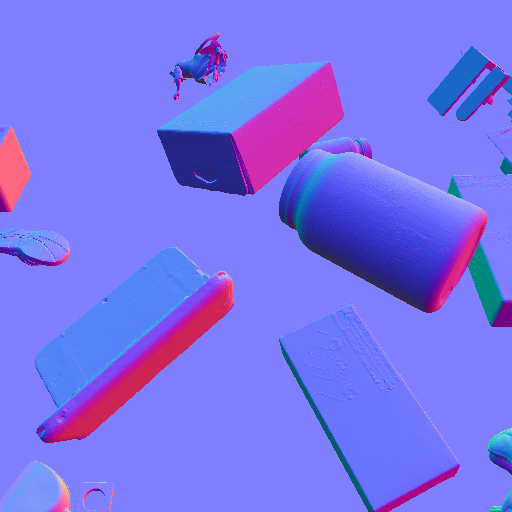} \\
        \rotatebox{90}{\textbf{(g)} Object coordinates}
        &\includegraphics[clip,trim={0cm 0cm 0cm 0cm},width=\linewidth]{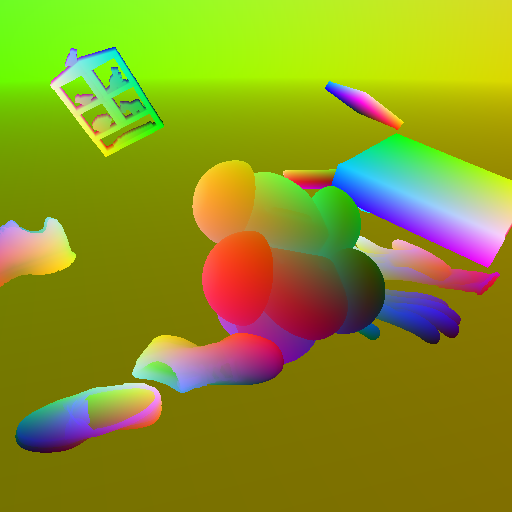}
        &\includegraphics[clip,trim={0cm 0cm 0cm 0cm},width=\linewidth]{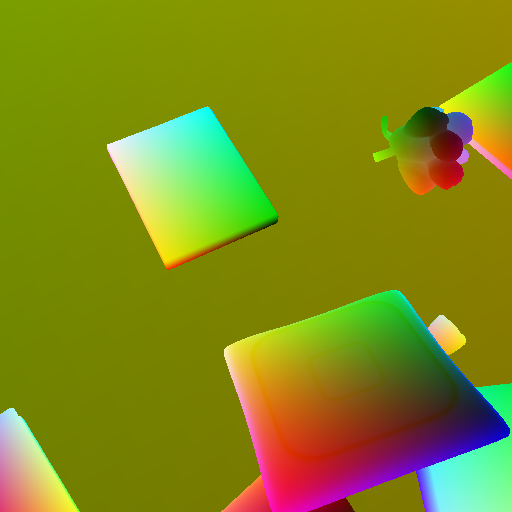}
        &\includegraphics[clip,trim={0cm 0cm 0cm 0cm},width=\linewidth]{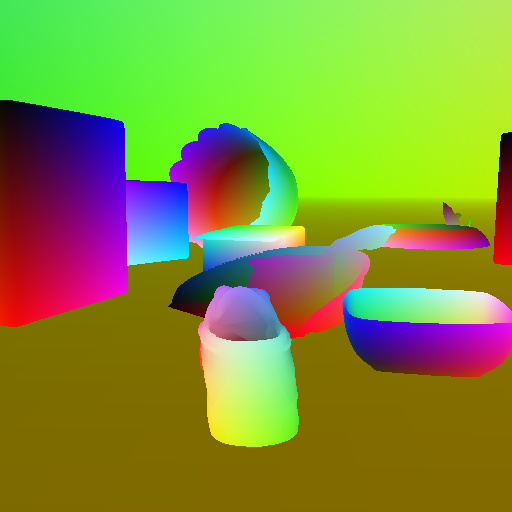}
        &\includegraphics[clip,trim={0cm 0cm 0cm 0cm},width=\linewidth]{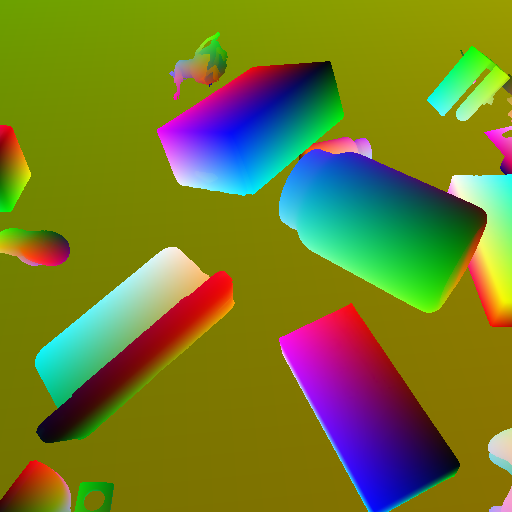} \\
        \rotatebox{90}{\textbf{(h)} Segmentation}
        &\includegraphics[clip,trim={0cm 0cm 0cm 0cm},width=\linewidth]{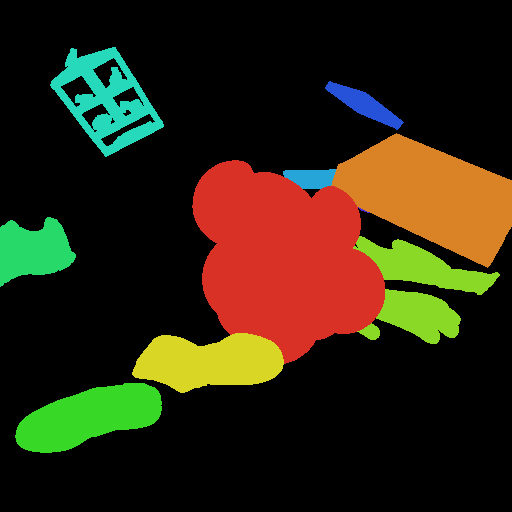}
        &\includegraphics[clip,trim={0cm 0cm 0cm 0cm},width=\linewidth]{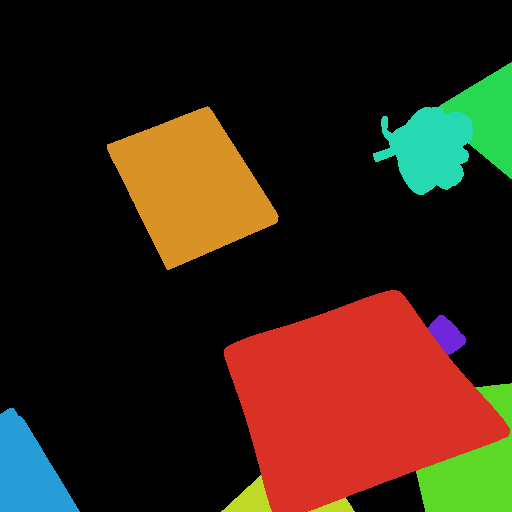}
        &\includegraphics[clip,trim={0cm 0cm 0cm 0cm},width=\linewidth]{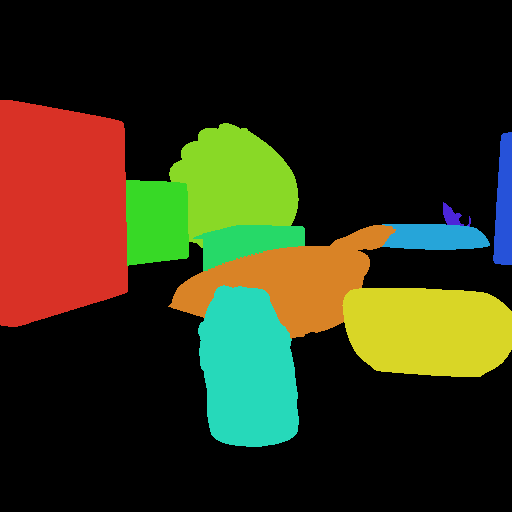}
        &\includegraphics[clip,trim={0cm 0cm 0cm 0cm},width=\linewidth]{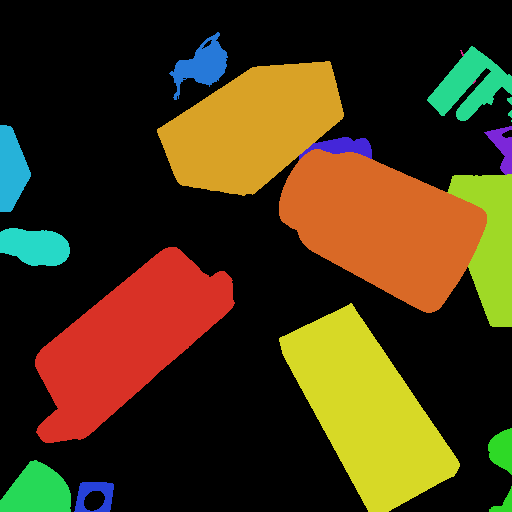} \\
    \end{tabular}
    }
    \caption{A few examples of {\dname}. Point tracks are subsampled for better visualization.}
    \label{fig:data:examples}
\end{figure*}

\def\figWidth{0.24\linewidth}
\begin{figure*}[t]
    \centering
    {\footnotesize
    \setlength{\tabcolsep}{1pt}
    \begin{tabular}{
    >{\centering\arraybackslash}m{0.26cm} 
    >{\centering\arraybackslash}m{\figWidth} 
    >{\centering\arraybackslash}m{\figWidth}
    >{\centering\arraybackslash}m{\figWidth}
    >{\centering\arraybackslash}m{\figWidth}
    }
        \rotatebox{90}{\makecell{GT}}
        &\gframe{\includegraphics[clip,trim={0 0 0 0},width=\linewidth]{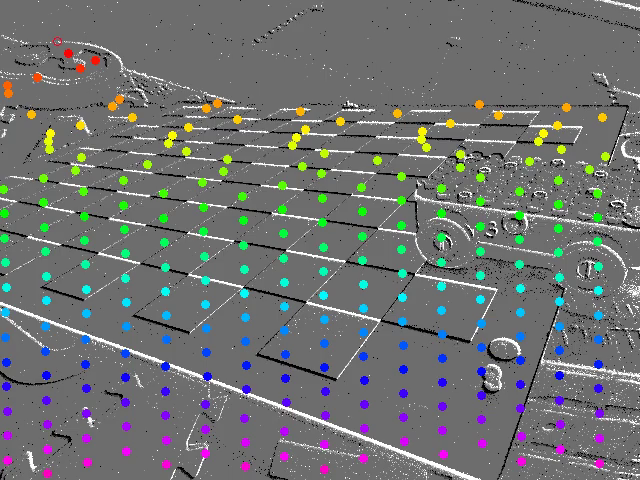}}
        &\gframe{\includegraphics[clip,trim={0 0 0 0},width=\linewidth]{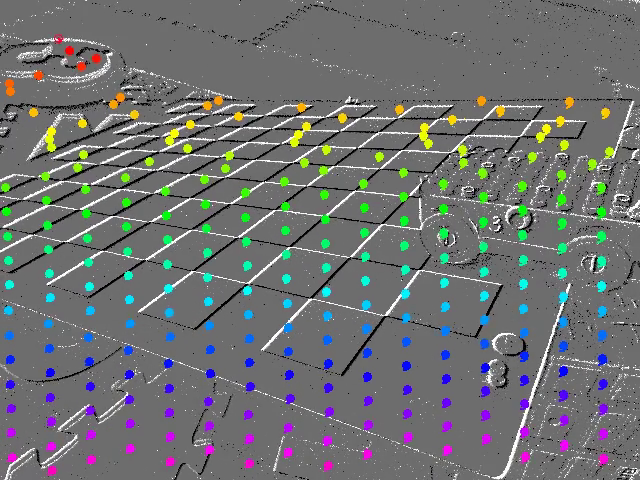}}
        &\gframe{\includegraphics[clip,trim={0 0 0 0},width=\linewidth]{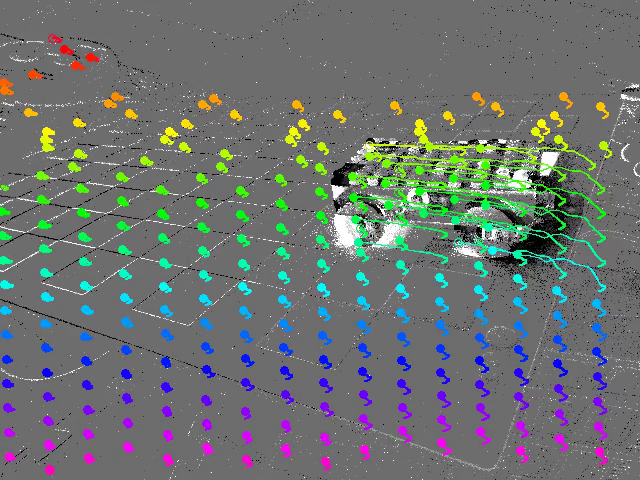}}
        &\gframe{\includegraphics[clip,trim={0 0 0 0},width=\linewidth]{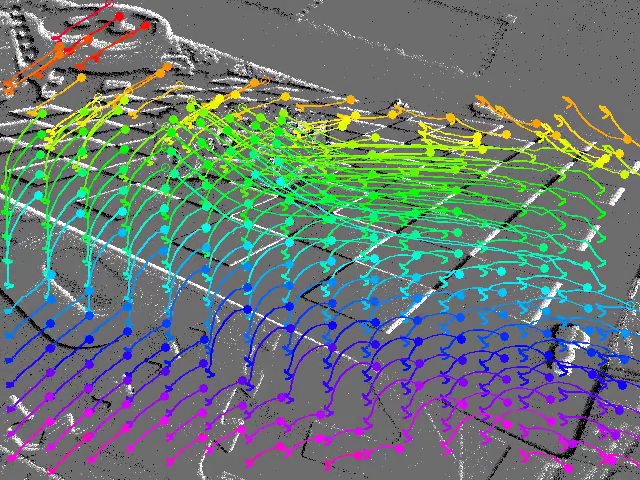}} \\
        \rotatebox{90}{\makecell{Ours}}
        &\gframe{\includegraphics[clip,trim={0 0 0 0},width=\linewidth]{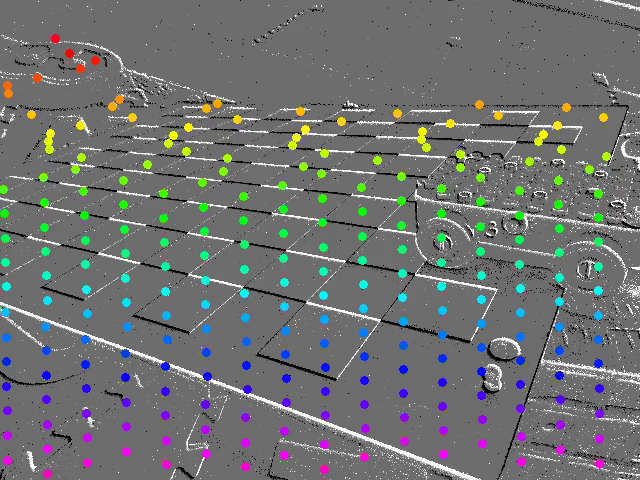}}
        &\gframe{\includegraphics[clip,trim={0 0 0 0},width=\linewidth]{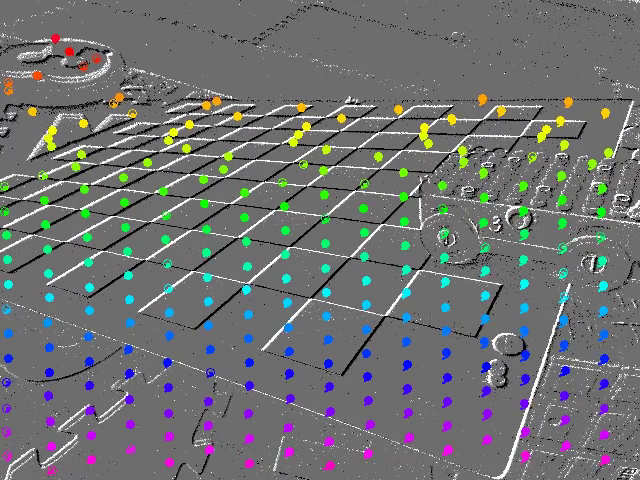}}
        &\gframe{\includegraphics[clip,trim={0 0 0 0},width=\linewidth]{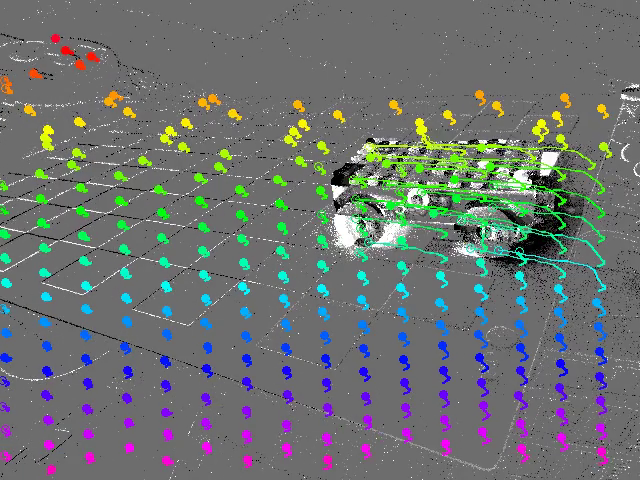}}
        &\gframe{\includegraphics[clip,trim={0 0 0 0},width=\linewidth]{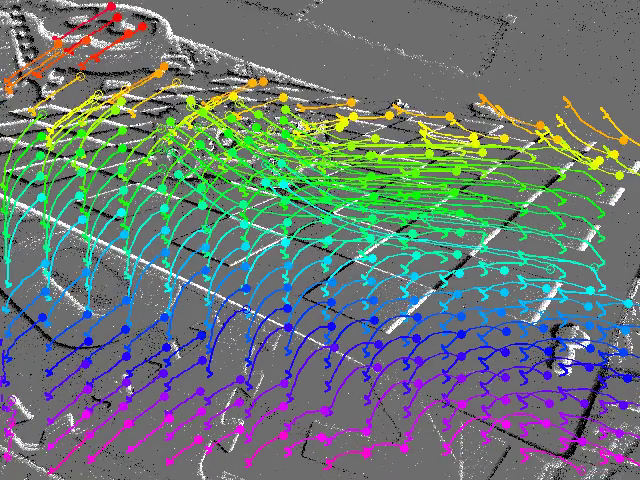}} \\[-0.5ex]
        & $t = 0s$
        & $t = 0.5s$
        & $t = 0.8s$
        & $t = 1.3s$ \\
        
        \rotatebox{90}{\makecell{GT}}
        &\gframe{\includegraphics[clip,trim={0 0 0 0},width=\linewidth]{images/fig_evimo_results/scene13_dyn_test_00_000000/gt/frame_00000000.png}}
        &\gframe{\includegraphics[clip,trim={0 0 0 0},width=\linewidth]{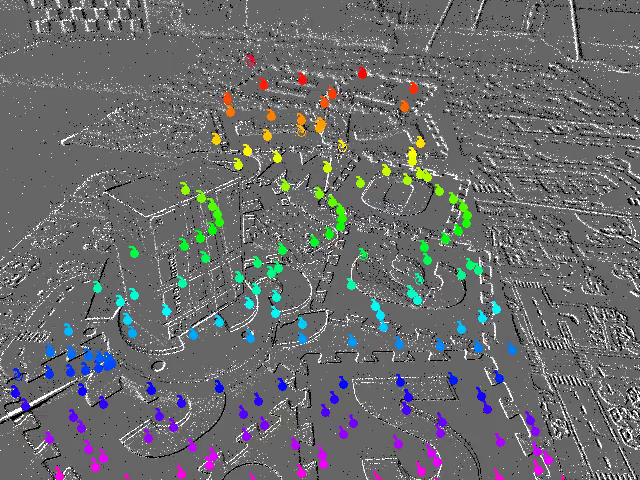}}
        &\gframe{\includegraphics[clip,trim={0 0 0 0},width=\linewidth]{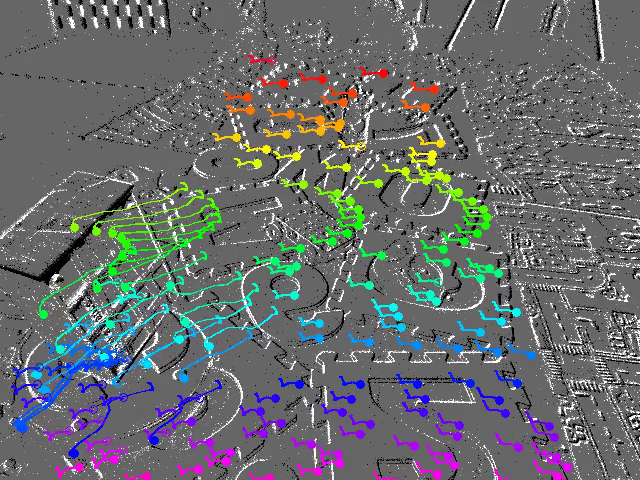}}
        &\gframe{\includegraphics[clip,trim={0 0 0 0},width=\linewidth]{images/fig_evimo_results/scene13_dyn_test_00_000000/gt/frame_00000095.png}} \\[1ex]
        \rotatebox{90}{\makecell{Ours}}
        &\gframe{\includegraphics[clip,trim={0 0 0 0},width=\linewidth]{images/fig_evimo_results/scene13_dyn_test_00_000000/pred/frame_00000000.png}}
        &\gframe{\includegraphics[clip,trim={0 0 0 0},width=\linewidth]{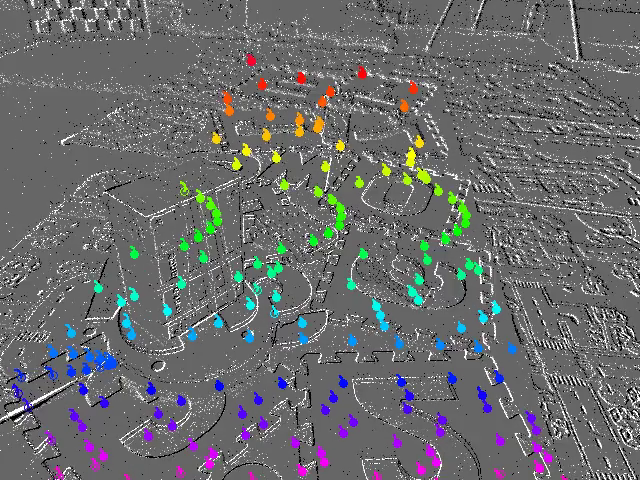}}
        &\gframe{\includegraphics[clip,trim={0 0 0 0},width=\linewidth]{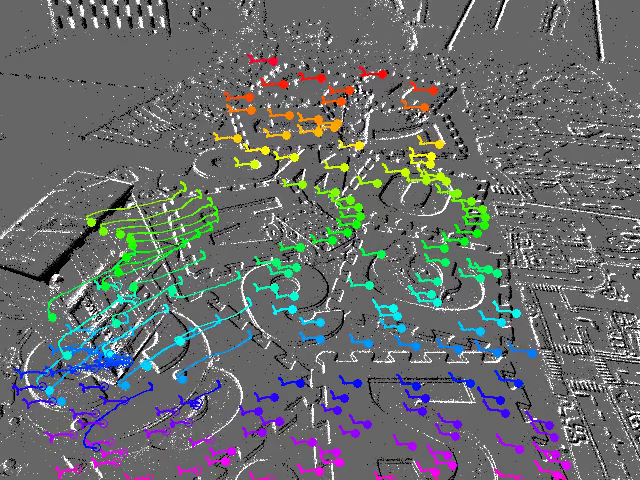}}
        &\gframe{\includegraphics[clip,trim={0 0 0 0},width=\linewidth]{images/fig_evimo_results/scene13_dyn_test_00_000000/pred/frame_00000095.png}} \\[-0.5ex]
        & $t = 0s$
        & $t = 0.57s$
        & $t = 0.97s$
        & $t = 1.5s$
    \end{tabular}
    }
\caption{\emph{Task 1 - TAP on EVIMO2 data}.
Visualization of track predictions.
}
\label{fig:evimo_pred}
\end{figure*}

\begin{table*}[t!]
\centering
\adjustbox{max width=\linewidth}{%
\setlength{\tabcolsep}{3pt}
\begin{tabular}{l c *{12}{S[table-format=1.3,detect-weight,detect-mode]}}
\toprule
  & & \multicolumn{2}{c}{Average} 
  & \multicolumn{2}{c}{Peanuts Light}
  & \multicolumn{2}{c}{Rocket Earth*}
  & \multicolumn{2}{c}{Ziggy Arena}
  & \multicolumn{2}{c}{Peanuts Running} \\
\cmidrule(l{1mm}r{1mm}){3-4}
\cmidrule(l{1mm}r{1mm}){5-6}
\cmidrule(l{1mm}r{1mm}){7-8}
\cmidrule(l{1mm}r{1mm}){9-10}
\cmidrule(l{1mm}r{1mm}){11-12}
Method & Frames & {FA$\uparrow$} & {EA$\uparrow$} 
       & {FA$\uparrow$} & {EA$\uparrow$}
       & {FA$\uparrow$} & {EA$\uparrow$}
       & {FA$\uparrow$} & {EA$\uparrow$}
       & {FA$\uparrow$} & {EA$\uparrow$} \\
\midrule
EKLT \cite{Gehrig19ijcv}                         & \cmark & 0.325 & 0.325 & 0.284 & 0.260 & 0.425 & 0.175 & 0.419 & 0.231 & 0.171 & 0.153 \\
DDFT \cite{Messikommer23cvpr}                    & \cmark & 0.576 & 0.472 & 0.447 & 0.420 & \unum{1.3}{0.648} & 0.291 & 0.748 & 0.746 & 0.460 & 0.428 \\
FE-TAP \cite{Liu24arxiv}                         & \cmark & 0.676 & 0.589 & \bnum{0.549} & \bnum{0.517} & 0.538 & 0.246 & \bnum{0.849} & \bnum{0.844} & \bnum{0.769} & \bnum{0.749} \\
\midrule
ICP \cite{Kueng16iros}                           & \xmark & 0.060 & 0.040 & 0.050 & 0.044 & 0.103 & 0.045 & 0.043 & 0.039 & 0.043 & 0.028 \\ 
EM-ICP \cite{Zhu17icra}                          & \xmark & 0.161 & 0.120 & 0.084 & 0.077 & 0.298 & 0.158 & 0.153 & 0.149 & 0.108 & 0.095 \\
HASTE \cite{Alzugaray20bmvc}                     & \xmark & 0.096 & 0.161 & 0.086 & 0.076 & 0.162 & 0.085 & 0.082 & 0.057 & 0.054 & 0.033 \\
DDFT E2VID \cite{Messikommer23cvpr}              & \xmark & 0.589 & 0.495 & {--}  & {--}  & {--}  &  {--} & {--}  & {--}  & {--}  & {--}  \\
\textbf{\mname~w\textbackslash o FA-loss~(Ours)} & \xmark & 0.698 & 0.599 & 0.538 & 0.508 & 0.676 & 0.336 & 0.842 & 0.841 & 0.736 & 0.713  \\
\textbf{\mname~(Ours)}                           & \xmark & 0.705 & 0.598 & 0.529 & 0.5   & 0.705 & 0.336 & 0.839 & 0.838 & 0.746 & 0.717 \\

\midrule \\[-1ex]
  & & \multicolumn{2}{c}{Average}
  & \multicolumn{2}{c}{shapes\_trans}
  & \multicolumn{2}{c}{shapes\_rot}
  & \multicolumn{2}{c}{shapes\_6dof}
  & \multicolumn{2}{c}{boxes\_trans}
  & \multicolumn{2}{c}{boxes\_rot} \\
\cmidrule(l{1mm}r{1mm}){3-4}
\cmidrule(l{1mm}r{1mm}){5-6}
\cmidrule(l{1mm}r{1mm}){7-8}
\cmidrule(l{1mm}r{1mm}){9-10}
\cmidrule(l{1mm}r{1mm}){11-12}
\cmidrule(l{1mm}r{1mm}){13-14}
Method & Frames & {FA$\uparrow$} & {EA$\uparrow$}
       & {FA$\uparrow$} & {EA$\uparrow$}
       & {FA$\uparrow$} & {EA$\uparrow$}
       & {FA$\uparrow$} & {EA$\uparrow$}
       & {FA$\uparrow$} & {EA$\uparrow$}
       & {FA$\uparrow$} & {EA$\uparrow$} \\
\midrule
EKLT \cite{Gehrig19ijcv}                            & \cmark & 0.811 & 0.775 & 0.839 & 0.740 & 0.833 & 0.806 & 0.817 & 0.696 & 0.682 & 0.644 & \unum{1.3}{0.883} & \unum{1.3}{0.865} \\
DDFT \cite{Messikommer23cvpr}                       & \cmark & 0.825 & 0.818 & 0.861 & 0.865 & 0.797 & 0.793 & 0.899 & 0.882 & \unum{1.3}{0.872} & \unum{1.3}{0.869} & 0.695 & 0.691 \\
FE-TAP \cite{Liu24arxiv}                            & \cmark & 0.844 & 0.838 & \bnum{0.931} & \bnum{0.929} & 0.815 & 0.813 & 0.879 & 0.860 & 0.731 & 0.728 & 0.862 & 0.861 \\
\midrule
ICP \cite{Kueng16iros}                              & \xmark & 0.256 & 0.245 & 0.307 & 0.306 & 0.341 & 0.339 & 0.169 & 0.129 & 0.268 & 0.261 & 0.191 & 0.188 \\
EM-ICP \cite{Zhu17icra}                             & \xmark & 0.337 & 0.334 & 0.403 & 0.402 & 0.320 & 0.320 & 0.248 & 0.242 & 0.355 & 0.354 & 0.356 & 0.349 \\
HASTE \cite{Alzugaray20bmvc}                        & \xmark & 0.442 & 0.427 & 0.589 & 0.564 & 0.613 & 0.582 & 0.133 & 0.043 & 0.382 & 0.368 & 0.492 & 0.447 \\
DDFT E2VID \cite{Messikommer23cvpr}                 & \xmark & 0.794 & 0.786 & {--}  & {--}  & {--}  & {--}  & {--}  & {--}  & {--}  & {--}  & {--}  & {--}  \\
\textbf{\mname~w\textbackslash o FA-loss~(Ours)}    & \xmark & 0.885 & 0.879 & 0.904 & 0.902 & 0.868 & 0.867 & 0.91  & 0.891 & 0.879 & 0.877 & 0.866 & 0.863  \\
\textbf{\mname~(Ours)}                              & \xmark & 0.888 & 0.883 & 0.91  & 0.904 & 0.867 & 0.865 & 0.904 & 0.886 & 0.866 & 0.864 & 0.896 & 0.893 \\

\bottomrule
\end{tabular}
}
\caption{Detailed performance comparison of tracking methods on the EDS (top) and EC (bottom) datasets.}
\label{tab:detailed-metrics}
\end{table*}
\def\figWidth{0.32\linewidth}
\begin{figure*}[ht!]
	\centering
    {\footnotesize
    \setlength{\tabcolsep}{2pt}
	\begin{tabular}{
	>{\centering\arraybackslash}m{0.3cm} 
	>{\centering\arraybackslash}m{\figWidth} 
	>{\centering\arraybackslash}m{\figWidth} 
	>{\centering\arraybackslash}m{\figWidth} 
    }

        \rotatebox{90}{\makecell{Ziggy in the Arena (EDS)}}
		&\gframe{\includegraphics[clip,trim={0cm 0cm 0cm 0cm},width=\linewidth]{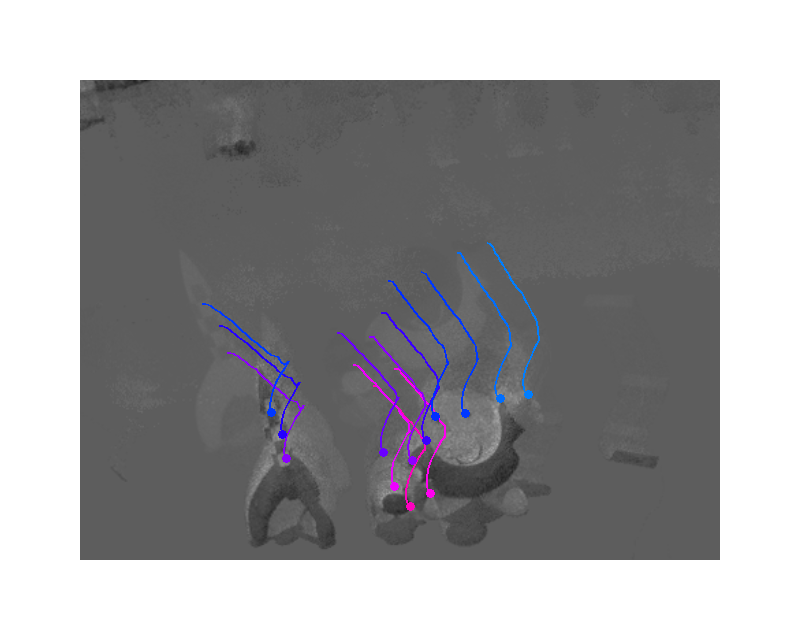}}
        &\gframe{\includegraphics[clip,trim={0cm 0cm 0cm 0cm},width=\linewidth]{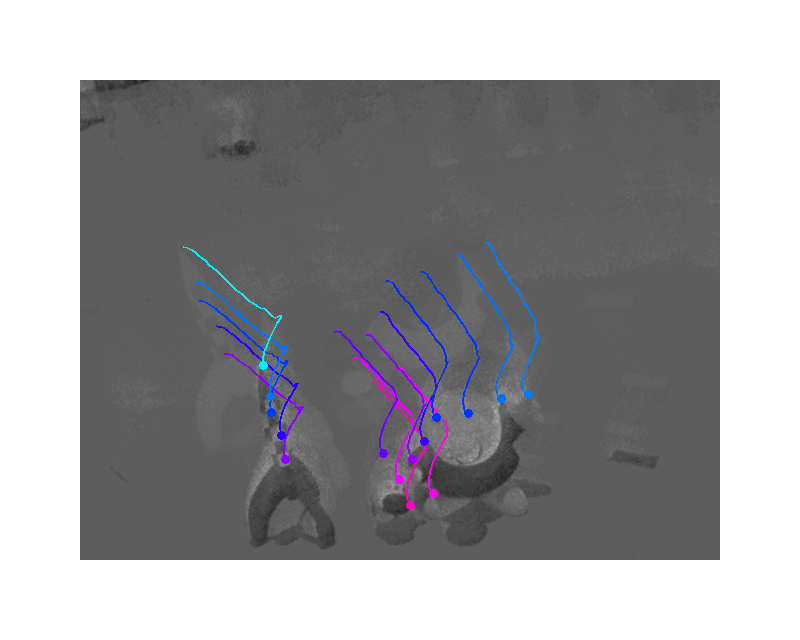}}
		&\gframe{\includegraphics[clip,trim={0cm 0cm 0cm 0cm},width=\linewidth]{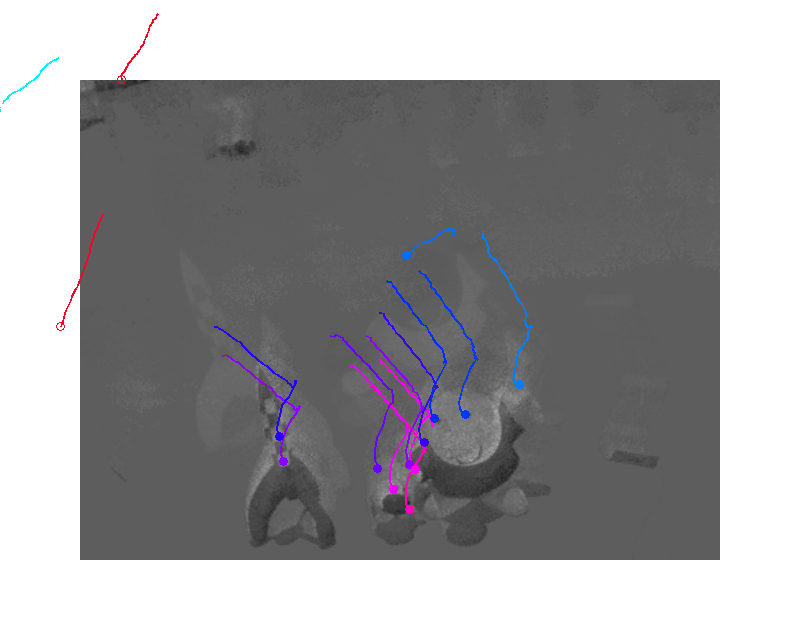}} \\

        \rotatebox{90}{\makecell{Peanuts Light (EDS)}}
		&\gframe{\includegraphics[clip,trim={0cm 0cm 0cm 0cm},width=\linewidth]{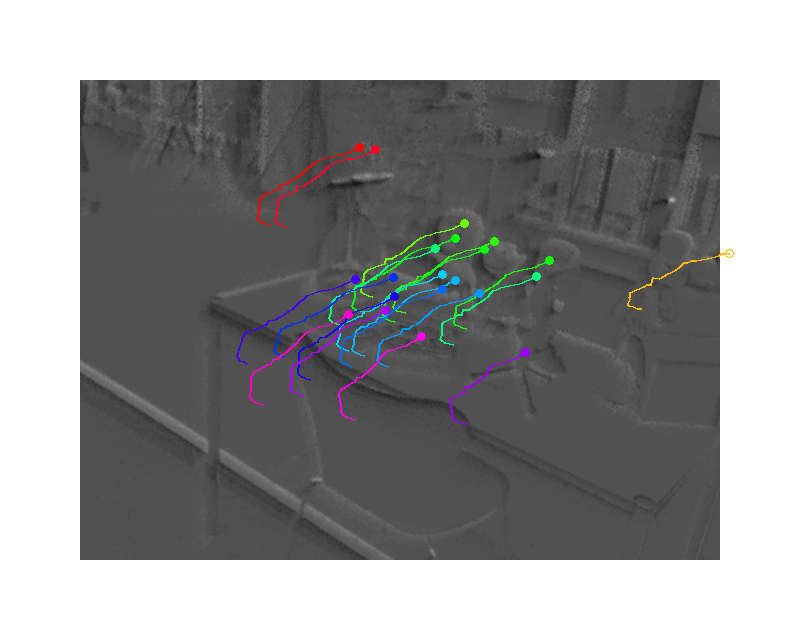}}
        &\gframe{\includegraphics[clip,trim={0cm 0cm 0cm 0cm},width=\linewidth]{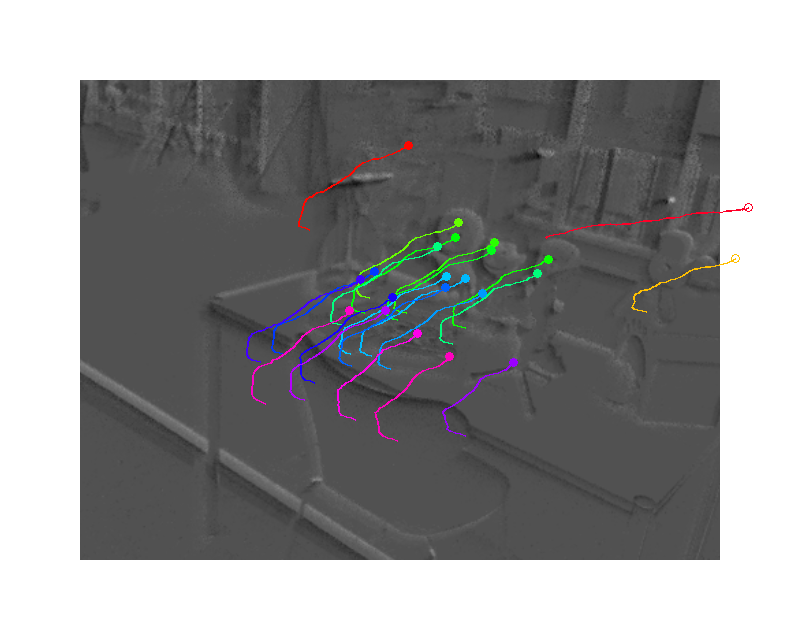}}
		&\gframe{\includegraphics[clip,trim={0cm 0cm 0cm 0cm},width=\linewidth]{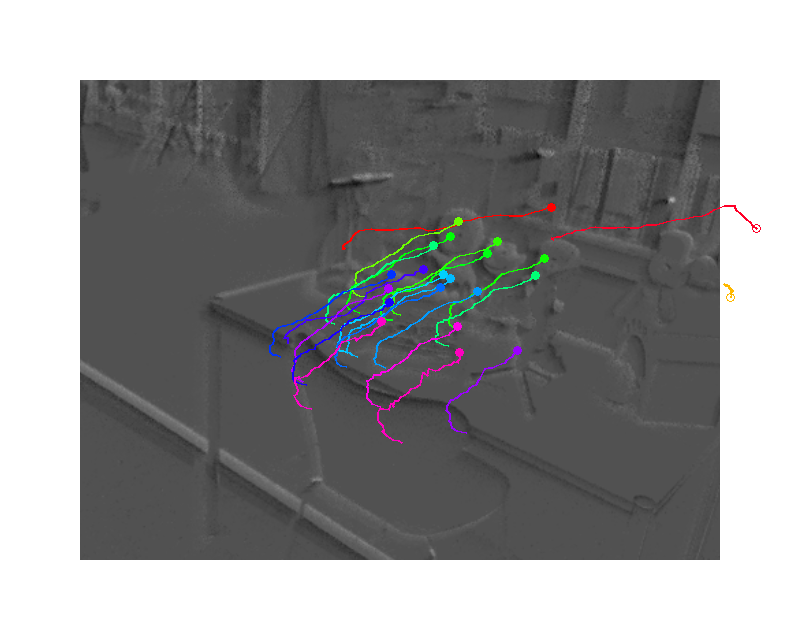}} \\

        \rotatebox{90}{\makecell{Shapes Rotation (EC)}}
		&\gframe{\includegraphics[clip,trim={0cm 0cm 0cm 0cm},width=\linewidth]{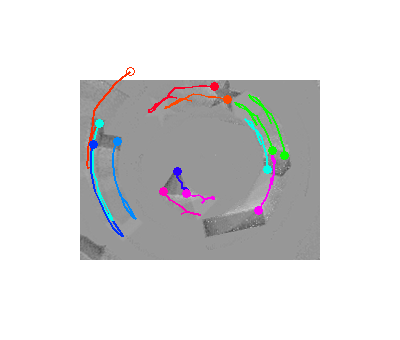}}
        &\gframe{\includegraphics[clip,trim={0cm 0cm 0cm 0cm},width=\linewidth]{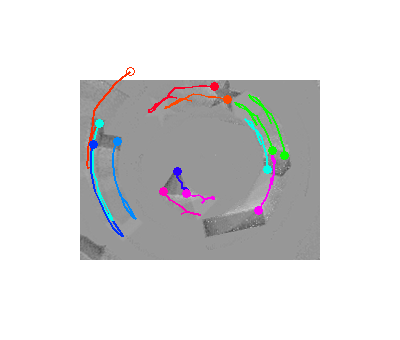}}
		&\gframe{\includegraphics[clip,trim={0cm 0cm 0cm 0cm},width=\linewidth]{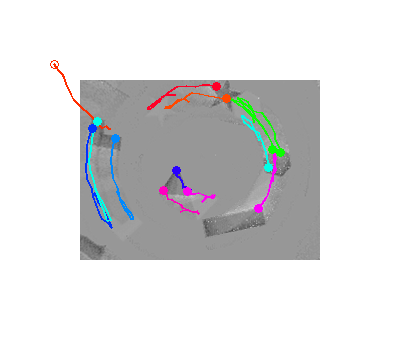}} \\

		& GT
		& \textbf{Ours}
		& DDFT~\cite{Messikommer23cvpr}
	\end{tabular}
	}
	\caption{Additional visualizations on the EDS and EC dataset.}
	\label{fig:exp:eds_ec_big}
\end{figure*}

\clearpage
{
    \small
    \bibliographystyle{ieeenat_fullname}
    \balance

}

\end{document}